\documentclass{article}

\usepackage{microtype}
\usepackage{graphicx}
\usepackage{subcaption}
\usepackage{booktabs} 

\usepackage{hyperref}

\usepackage[preprint]{icml2026}

\usepackage{amsmath}
\usepackage{amssymb}
\usepackage{mathtools}
\usepackage{amsthm}

\usepackage[capitalize,noabbrev]{cleveref}

\theoremstyle{plain}

\theoremstyle{definition}

\theoremstyle{remark}

\usepackage[textsize=tiny]{todonotes}

\icmltitlerunning{StAD: Stein Amortized Divergence}

\DeclareMathOperator*{\argmin}{arg\,min}

\begin{document}

\twocolumn[
  \icmltitle{StAD: Stein Amortized Divergence \\ for Fast Likelihoods with Diffusion and Flow}



  \icmlsetsymbol{equal}{*}

  \begin{icmlauthorlist}
    \icmlauthor{Gurjeet Jagwani}{ioa,rcs}
    \icmlauthor{Stephen Thorp}{ioa}
    \icmlauthor{Sinan Deger}{ioa}
    \icmlauthor{Hiranya Peiris}{ioa,cav,okc}
  \end{icmlauthorlist}

  \icmlaffiliation{ioa}{Institute of Astronomy and Kavli Institute for Cosmology, University of Cambridge, Cambridge, UK}
  \icmlaffiliation{rcs}{Research Computing Services, University of Cambridge, Cambridge, UK}
  \icmlaffiliation{okc}{The Oskar Klein Centre, Department of Physics, Stockholm University, Stockholm, SE}
  \icmlaffiliation{cav}{Cavendish Laboratory, Department of Physics, University of Cambridge, Cambridge, UK}

  \icmlcorrespondingauthor{Gurjeet Jagwani}{gurjeet.jagwani@ast.cam.ac.uk}

  \icmlkeywords{Machine Learning, ICML}

  \vskip 0.3in
]



\printAffiliationsAndNotice{}  

\newcommand{\E}{\mathbb{E}}
\newcommand{\R}{\mathbb{R}}
\newcommand{\dd}{\mathrm{d}}
\newcommand{\score}{\mathbf{s}}
\newcommand{\vv}{\mathbf{v}}
\newcommand{\xx}{\mathbf{x}}
\newcommand{\grad}{\nabla}
\newcommand{\divergence}{\nabla\!\cdot}
\newcommand{\inner}[2]{\big\langle #1,\, #2 \big\rangle}

\begin{abstract}
Diffusion and flow-based models are ubiquitously used for generative modelling and density estimation. They admit a deterministic probability flow ordinary differential equation (PF-ODE), analogous to continuous normalizing flows (CNFs), which describes the transport of the probability mass. Obtaining the likelihood from these models is of interest to many workflows, especially Bayesian analysis, and requires solving the trace of the Jacobian to compute the divergence of the learned PF-ODE, which is either $\mathcal{O}(D^2)$ to compute exactly or $\mathcal{O}(D)$ 
 with a noisy estimate. We introduce \textit{StAD}, a new distillation method to predict and learn the divergence of the PF-ODE using the Langevin-Stein operator without ever computing the Jacobian. We show that our method is competitive with the Hutchinson and Hutch++ on CIFAR-10, ImageNet and other density estimation tasks, consistently improving the variance and speed of the likelihood predictions compared to the Hutchinson. We additionally show our method will generalize to a varied class of generative models, and show that under some regularity conditions these learned vector fields can be made to satisfy the Stein class.
\end{abstract}

\section{Introduction}
Density estimation is of great interest to many domains like molecular dynamics, neuroscience, cosmology, astrophysics, materials science, evolutionary biology, economics, simulation-based inference, and reinforcement learning among others \citep[e.g.][]{arnaudon2016stochastic, noé2019boltzmanngeneratorssampling, Alsing_2019, schebek2025boltzmann}.
Diffusion \citep{sohl2015diffusion, song2019gradients, song2020score, ho2020diffusion, song2021sde, song2021likelihood} and flow-matching \citep{lipman2023flowmatchinggenerativemodeling, lipman2024flowmatching} models are the current state-of-the-art algorithms for estimating the high-dimensional densities commonly encountered within many scientific and non-scientific datasets \citep[for a review, see][]{yang2023diffusionreview, arruda2025diffusionsbi}. They fulfil a similar role as other probabilistic models such as normalizing flows \citep[reviewed in][]{kobyzev2021nf, papamakarios2021nf}, and build on earlier pioneering work on score-based models \citep[e.g.][]{hyvarinen2005score, vincent2011score}. 

\subsection{Divergence under diffusion and flow}

\begin{figure}[ht]
  \begin{center}
    \centerline{\includegraphics[width=\columnwidth]{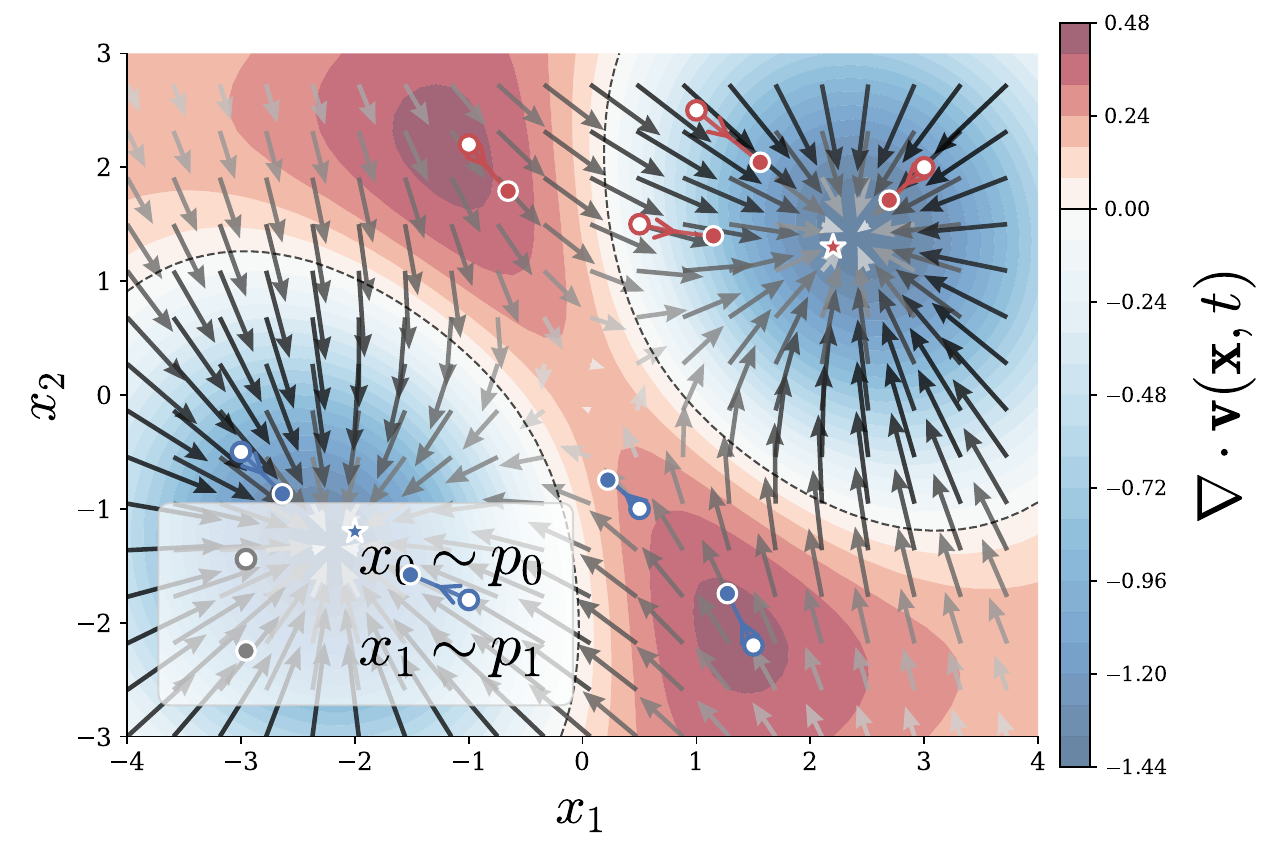}}
    \caption{
    Divergence (blue/red contours) visualisation for a vector field (arrows) defined by the PF-ODE at a fixed time step.
    }
    \label{img:divergencevisualisation}
  \end{center}
\end{figure}

Diffusion and flow-based models are concerned with learning a probability density $p(\xx)$ over a variable $\xx$ in $\R^D$. This is represented as a continuous-time transform between the target density $p_\varepsilon(\xx_\varepsilon)$ at time $t=\varepsilon\gtrsim0$, and a simple base (noise) density $p_T(\xx_T)$ at time $t=T$. Diffusion models learn the score $\score(\xx_t)=\grad\log p(\xx_t)$ of the changing probability density function, with a stochastic differential equation (SDE) guiding this process \citep{song2021sde}. Flow-matching models learn a conditional vector field that is analogous to the marginal vector field describing the transport of the probability mass from noise to data \citep{lipman2023flowmatchinggenerativemodeling}. They both can be interpreted as continuous normalising flows (CNFs; \citealp{chen2018node, grathwohl2019ffjord}), or a probability flow ordinary differential equation (PF-ODE; \citealp{song2021sde}). For both classes of models, the PF-ODE has a general form:
\begin{equation}
    \frac{\dd\xx_t}{\dd t} = \vv_t(\xx_t), \qquad t\in[\varepsilon,T],
    \label{eq:pf-ode}
\end{equation}
where $\vv_t$ is any vector field or differential equation describing the transport of probability. In diffusion models---obeying an It\^o SDE of the form $\dd\xx_t=\mathbf{f}(\xx_t,t)\dd t+g(t)\dd\mathbf{w}$, where $\mathbf{f}$ and $g$ are \emph{drift} and \emph{diffusion} coefficients, and $\mathbf{w}$ follows a Weiner process---the PF-ODE has the form $\vv_t(\xx_t)=\mathbf{f}(\xx_t,t) - \frac{1}{2}g(t)^2\,\score (\xx_t)$ \citep[see][]{maoutsa2020fokker, song2021sde}. Flow-matching models admit a simulation-free objective to directly approximate $\vv_t$.
The instantaneous change-of-variables formula from CNFs \citep{chen2018node, grathwohl2019ffjord} gives
\begin{equation}
    \frac{\dd}{\dd t}\log p_t(\xx_t) \;=\; -\,\inner{\nabla}{\vv_t(\xx_t)}.
    \label{eq:icov}
\end{equation}
Integrating \eqref{eq:icov} for $t=T$ to $t=\varepsilon$ \citep[as in][]{song2021sde} allows one to compute the log probability (likelihood) of a variable under the target density,
\begin{equation}
    \log p_\varepsilon(\xx_\varepsilon) = \log p_T(\xx_T) + \int^{T}_{\varepsilon} \inner{\nabla}{\vv_t(\xx_t)}\,\dd t, \label{eq:lik}
\end{equation}
using an initial value $\xx_t=\xx_\varepsilon$ at $t=\varepsilon$, and simultaneously solving \eqref{eq:pf-ode} for $\xx_T$. The bottleneck here is the divergence term---normally computed using the trace of the Jacobian $\inner{\nabla}{\vv_t(\xx_t)}=\text{Tr} ( \partial \vv_t/\partial \xx_t)$---which is of the time complexity $\mathcal{O}(D^2)$ to determine exactly, or $\mathcal{O}(D)$ to approximate using noisy unbiased trace estimators like Hutchinson \citep{hutchinson1989trace, skilling1989trace, girard1989montecarlo, grathwohl2019ffjord} or Hutch++ \cite{meyer2021hutchoptimalstochastictrace, liu2025trace}. Fig.~\ref{img:divergencevisualisation} illustrates an example PF-ODE vector field, and the corresponding divergence scalar field, at a single time step.

\subsection{Stein's method and Stein operators}
Stein's method \cite{stein1972bound} was introduced to bound the distance between a sum of random variables and the Gaussian distribution. Since then, it has proved to be a powerful tool for quantifying distances between probability measures and bounding convergence rates in limit theorems \cite{Barbour_1988}, and has been extended considerably beyond its original application \citep[for a review, see][]{anastasiou2023stein}. Stein operators $\mathcal{A}_p$ are defined as linear operators acting on a set of test functions $\mathcal{U}$, that follow the Stein identity,
\begin{equation}
    \E_{x \sim p}[(\mathcal{A}_pu)(x)] = 0, \qquad \forall  u\in \mathcal{U},
\end{equation}
where $x$ is a random variable from a target distribution $p$. 

The Langevin--Stein operator is an extension of this formalism to $\R^D$. For any density $p(\xx)$ with score $\score(\xx)= \nabla \log p(\xx)$ and any vector field $\vv(\xx)$ that is in the Stein class, the Langevin--Stein operator,
\begin{equation}
    \mathcal{A}_p \vv(\xx) \;=\; \inner{\nabla}{\vv(\xx)} + \inner{\vv(\xx)}{\score(\xx)},
    \label{eq:langevin-stein}
\end{equation}
satisfies the Stein identity, 
\begin{equation}
    \mathbb{E}_{\xx\sim p}[\mathcal{A}_p \vv(\xx)]=0,
    \label{eq:stein-identity}
\end{equation} 
under mild boundary conditions \citep{gorham2015, gorham2017ksd, liu2016ksd}; see further details in Appendix \ref{sec:stein-class} and \S\ref{sec:regularisation}. Generalizations of the Langevin--Stein operator exist to encompass broader classes of target function (e.g.\ the diffusion Stein operator; \citealp{gorham2019}).

\subsection{Outline and contributions}
In this work, we show that diffusion and flow-matching models satisfy these boundary conditions with some mild assumptions and derive a novel method---the \textit{Stein amortized divergence} (\textit{StAD})---that can learn the divergence of the PF-ODE using the Langevin-Stein operator (\ref{eq:langevin-stein}). The learned network can then predict the divergence of the vector field at any given time step and produce likelihoods with $\mathcal{O}(D)$ time complexity and lower variance than the Hutchinson and Hutch++ estimators.

In \S\ref{sec:related}, we review related methods, including general uses of Stein's method, other trace estimators, and the distillation of diffusion/flow models. In \S\ref{sec:methodology}, we derive \textit{StAD} for diffusion and flow-matching. Additionally, we describe regularisation methods for stable training and satisfying the Stein class. In \S\ref{sec:experiments} we show results compared to the Hutchinson and Hutch++ estimators for a range of density estimation problems. We conclude the paper in \S\ref{sec:conclusion}.

\section{Related work}
\label{sec:related}
\subsection{Recent applications of Stein's method}
Stein's method has been widely adapted as a distance measure between distributions in computational statistics \citep[e.g.][]{liu2016ksd, barp2019msd}. It has also been utilised to test the quality of a sample generated from a target distribution under MCMC and variational inference settings \citep{gorham2015, gorham2017ksd, gorham2019}. \citet{chen2018steinpoints} have further developed methodology to deterministically select data points that yield an accurate approximate representation of the posterior, by minimising the Kernel Stein Discrepancy (KSD; \citealp{chwialkowski2016kernel}). Stein thinning has also been introduced to post-process MCMC outputs and correct for bias \citep{10.1111/rssb.12503, bénard2023kernelsteindiscrepancythinning}. Additionally, \citet{oates2016controlfunctionalsmontecarlo} design control variates for Monte Carlo estimators using the Stein identity. Recent work has been done to introduce Stein's method into more modern machine learning frameworks. \citet{liu2016svgd} introduce Stein Variational Gradient Descent (SVGD) which allows for Bayesian inference as an optimisation problem using Stein discrepancy (see also \citealp{ranganath2016operator}).  More recent work by \citet{grathwohl2020learningsteindiscrepancytraining} shows how Stein discrepancies can be utilised to train and evaluate energy-based models. 

\subsection{Trace estimators and tractable Jacobians}
As established before, to compute the divergence of the relevant PF-ODE, we need to compute the trace of the Jacobian. Common approaches to achieve this are the Hutchinson \citep{hutchinson1989trace, skilling1989trace, girard1989montecarlo, silber1994megadimensional, grathwohl2019ffjord} and Hutch++ estimators \citep{meyer2021hutchoptimalstochastictrace, liu2025trace}. Hutchinson introduced a stochastic method to estimate the trace of a $D\times D$ matrix $\mathbf{A}$ using random noise vectors or \textit{probes}, $\mathbf{n}\in\R^D$, that are Rademacher or Gaussian distributed (for a review, see e.g.\ \citealp{avron2011trace, roosta2015trace, lin2016spectral, adams2018spectral, jiang2021optimalsketching}). The trace of $\mathbf{A}$ is then estimated in an unbiased way by averaging over probes: $\text{Tr}(\mathbf{A}) \approx \E_\mathbf{n}[\mathbf{n}^\top\mathbf{A}\mathbf{n}]$. The Hutch++ algorithm introduced by \citet{meyer2021hutchoptimalstochastictrace} provides comparable performance for positive semi-definite (PSD) matrices with fewer matrix--vector multiplications by using a low-rank projection (see also the work of \citealp{lin2017spectral, saibaba2017trace, gambhir2017deflation, li2021krylov}). An $n$-probe, rank-$n$ implementation of Hutch++ proceeds by first generating random matrices $\mathbf{S},\mathbf{G}\in\R^{D\times n}$ each with Rademacher-distributed elements. This is followed by a $QR$ decomposition, $\mathbf{Q}\mathbf{R}=\mathbf{A}\mathbf{S}$, yielding an orthonormal basis $\mathbf{Q}\in\R^{D\times n}$. The trace of $\mathbf{A}$ can then be approximated by
\begin{multline}
    \text{Tr}(\mathbf{A}) \approx \text{Tr}(\mathbf{Q}^\top\mathbf{A}\mathbf{Q}) \\+ \frac{1}{n}\text{Tr}\Big(\mathbf{G}^\top(\mathbf{I}-\mathbf{QQ}^\top)\mathbf{A}(\mathbf{I-\mathbf{QQ}^\top})\mathbf{G}\Big),
\end{multline}
requiring $3n$ matrix--vector multiplications.

\citet{persson2022hutch} presented an adaptive variant of Hutch++, building on the work of \citet{meyer2021hutchoptimalstochastictrace}, \citet{jiang2021optimalsketching}, and \citet{gittens2013nystrom}. XTrace and XNysTrace, introduced by \citet{Epperly_2024}, are the latest advancement in trace estimators. These are based on a sum over basic trace estimators, achieving much smaller errors for the same matrix--vector multiplication budget as Hutch++. The XTrace algorithm is also based on Rademacher-distributed random vectors $\mathbf{\Omega}\in\R^{D\times n}$ and a $QR$ decomposition $\mathbf{Q}\mathbf{R}=\mathbf{A}\mathbf{\Omega}$, yielding $\mathbf{Q}\in\R^{D\times n}$, $\mathbf{R}\in\R^{n\times n}$ for $n\leq D$. \citet{Epperly_2024} derive an efficient trace estimator as an average over $n$ individual estimators, each depending on $\mathbf{Q}^\top\mathbf{A}\mathbf{Q}$, and rows and columns of $\mathbf{R}^{-1}$, $\mathbf{Q}^\top\mathbf{\Omega}$, and $(\mathbf{A}\mathbf{Q})^\top\mathbf{\Omega}$. The resulting estimator requires $2n$ matrix--vector products. Appendix \ref{sec:hutch-comparison} performs a comparison of Hutch++ and XTrace for random matrices.

\citet{grathwohl2019ffjord}, establish how the Hutchinson can be used to train CNFs with cheap likelihood estimates, and allows for more expressive architectures than the predecessors. \citet{liu2025trace} again builds upon this, and replaces the Hutchinson with the Hutch++ estimator, and attempts to make it scaleable by caching the $\mathbf{Q}$ matrix during computations.

A parallel direction is the development of models with Jacobians that are engineered for tractable trace or determinant estimation. \citet{dinh2015nicenonlinearindependentcomponents} design non-linear components for estimating the relevant density, and construct Jacobians that have a determinant that is trivial to compute. \citet{chen2019differential} propose volume-preserving flows by extending \citet{dinh2015nicenonlinearindependentcomponents} to continuous flows that would have $\inner{\nabla}{\vv} = 0$ by construction \citep[see also][]{toth2020hamiltonian, bilos2021volume}. 

\subsection{Distillation of diffusion and flow-based models}
Diffusion and flow-based models generally solve a neural ODE to transform the initial noise to the target distribution. Solving this system is expensive and requires many steps or evaluations along the trajectory. To enable fast sampling, many distillation techniques have been introduced. \citet{salimans2022progressivedistillationfastsampling} introduced progressive distillation to teach a student network to solve the trajectory with less time-steps, \citet{song2023consistency} introduce consistency models which make time jumps using time-discretization solvers, and \citet{lu2025simplifying} stabilise them by making them follow trigonometric identities. It was observed that having straight paths leads to faster ODE solves, leading to the development of the ReFlow framework \citep{liu2023flow, liu2024instaflowstephighqualitydiffusionbased}. Furthermore, score identity distillation techniques, were introduced for diffusion models \citep{zhou2024scoreidentitydistillationexponentially} and were recently extended to flow-matching \citep{zhou2025scoredistillationflowmatching}. Other distillation methods for few and 1-step generation have shown that it is relatively easy to learn a student network to reduce inference costs of the teacher \citep[e.g.][]{luhman2021knowledgedistillationiterativegenerative, luo2023diffinstruct, yin2024onestepdiffusiondistributionmatching,  sauer2023adversarialdiffusiondistillation}.
 More recently, \citet{ai2025distillation} introduced a joint-distillation framework, which can learn the divergence of the continuous neural-ODE using the Hutchinson estimator (for a direct comparison see Appendix \ref{sec:direct_divergence}). This allows for fast sampling and few step generation in addition to accurate likelihoods. Similarly, \citet{rehman2025falconfewstepaccuratelikelihoods}, construct few-step generators with accurate enough likelihoods for importance sampling.

\section{Methodology}
\label{sec:methodology}
The Langevin--Stein operator \eqref{eq:langevin-stein}, defines the relationship between the divergence $\inner{\nabla}{ \vv}$, the vector field $\vv$ and the score $\score$, in terms of expectations,
\begin{equation}
    \E_{\xx}\Big[\inner{\nabla}{\vv(\xx)} + \inner{\vv(\xx)}{\score(\xx)}\Big] = 0.
    \label{eq:langevin-stein-E}
\end{equation}
The operator can thus be used to define the average divergence of the vector field,
\begin{equation}
     \E_{\xx}\big[\inner{\nabla}{\vv(\xx)}\big] = - \E_{\xx}\Big[\inner{\vv(\xx)}{\score(\xx)}\Big],
\end{equation}
which can be used to define a pointwise Stein baseline,
\begin{equation}
    \label{eq: baseline}
    b(\xx) \coloneqq -\inner{\vv(\xx)}{\score(\xx)},
\end{equation}
which under the expectation has the property $\E_{\xx\sim p}[b(\xx)]=\E_{\xx\sim p}[\inner{\nabla}{ \vv(\xx)}]$. Since we already have an estimate of $\vv(\xx)$ and $\score(\xx)$ with score-based models, this is trivial to compute. It has been shown that the score of flow-matching models can be easily obtained under Gaussian assumptions \citep[e.g.][]{zheng2023guided, lipman2024flowmatching, zhou2025scoredistillationflowmatching}; this again allows us to define and compute a cheap baseline that captures the average $\inner{\nabla}{ \vv(\xx)}$. To get a more accurate pointwise estimate, we learn a correction term $\delta(\xx)$ at each time step. We can define the pointwise divergence as
\begin{equation}
    \inner{\nabla}{ \vv(\xx)} = b(\xx) + r(\xx),
    \label{eq:pointwise-divergence}
\end{equation}
where $r(\xx)$ is the target residual we want to learn with $\delta(\xx)$.

The most direct objective to learn an optimal estimate, $\delta^*(\xx)$, of the divergence residual, $r(\xx)$, would be the $L^2$ regression:
\begin{equation}
    \delta^*(\xx) = \argmin_{\delta}\;\E_{\xx}\Big[\big(\delta(\xx)-r(\xx)\big)^2\Big].
    \label{eq:ideal}
\end{equation} 
Here the target residual,
\begin{equation}
\label{eq:residual}
    r(\xx):= \inner{\nabla}{\vv(\xx)}+ \inner{\vv(\xx)}{\score(\xx)} = \mathcal{A}_p \vv(\xx),
\end{equation}
is unknown because it contains $\inner{\nabla}{ \vv(\xx)}$.

Expanding \eqref{eq:ideal} with the binomial theorem,
\begin{multline}
    \E_{\xx}\big[(\delta(\xx)-r(\xx))^2\big] \\= \E_\xx\big[\delta^2(\xx)\big]-2\,\E_\xx\big[\delta(\xx)\,r(\xx)\big]+\E_\xx\big[r^2(\xx)\big].
    \label{eq:binomial}
\end{multline}
By the Stein identity and divergence theorem, the mixed term in \eqref{eq:binomial} can be rewritten \emph{without} a dependence on $\inner{\nabla}{ \vv}$:
\begin{equation}
    \E_{\xx}\!\big[\delta(\xx)\,r(\xx)\big] \;=\; -\,\E_{\xx}\!\Big[\inner{\grad \delta(\xx)}{\vv(\xx)}\Big].
\end{equation}

We can show this by rewriting the mixed term as an expectation integral, and using the fact that $\score(\xx)=\nabla\log p(\xx) = \big(\nabla p(\xx)\big)/p(\xx)$, which gives
\begin{equation}
    \begin{split}
        &\E_{\xx}\big[\delta(\xx)\,r(\xx)\big]\\
        &= \int_{\R^D}\big[p(\xx)\,\delta(\xx)\,\inner{\nabla}{\vv(\xx)} + \delta(\xx)\,\inner{\vv(\xx)}{\nabla p(\xx)}\big]\,\dd \xx.
      \end{split}
      \label{eq:mixedintegral}
\end{equation}
Using the product rule, we can rewrite \eqref{eq:mixedintegral} as
\begin{multline}
    \E_{\xx}\big[\delta(\xx)\,r(\xx)\big] = \int_{\mathbb{R}^D}\inner{\nabla}{p(\xx)\,\delta(\xx)\,\vv(\xx)}\,\dd \xx \\
    - \int_{\mathbb{R}^D} p(\xx)\,\inner{\nabla\delta(\xx)}{\vv(\xx)}\,\dd \xx.
    \label{eq:mixedintegraldiv}
\end{multline}

To interpret the first term in \eqref{eq:mixedintegraldiv}, we can apply the divergence theorem on the ball $B_R=\{\xx\in\R^D:\|\xx\|\le R\}$, and then take $R\to\infty$. The divergence integral becomes
\begin{multline}
    \int_{B_R}\inner{\nabla}{p(\xx)\,\delta(\xx)\,\vv(\xx)}\,\dd \xx \\
    = \oint_{\partial B_R} \inner{p(\xx)\,\delta(\xx)\,\vv(\xx)}{ \hat{\mathbf{n}}}\,\dd S,
    \label{eq:vanishingmarginal}
\end{multline}
where $\hat{\mathbf{n}}\coloneq\xx/||\xx||$ is the outward unit normal on $\partial B_R$,
and  and $\dd S$ is the $(D-1)$-dimensional surface measure. For diffusion marginals at $t\geq\varepsilon$, and vector fields of at most polynomial growth, $p(\xx)\,\delta(\xx)\,\vv(\xx)$ decays fast enough that the boundary term vanishes~\citep{liu2016ksd, liu2016svgd,gorham2017ksd,barp2019msd}; i.e.,
\begin{equation}
\lim_{R\to\infty}\oint_{\partial B_R} \inner{p(\xx)\,\delta(\xx)\,\vv(\xx)}{ \hat{\mathbf{n}}}\,\dd S = 0.
\end{equation}
Then we can simplify \eqref{eq:mixedintegraldiv} to
\begin{equation}
            \E_{\xx}\big[\delta(\xx)\,r(\xx)\big]
        = -\E_{\xx}\Big[\inner{\nabla\delta(\xx)}{\vv(\xx)}\Big].
\end{equation}

Plugging this into \eqref{eq:binomial} shows that the optimization in \eqref{eq:ideal} is equivalent---up to the constant $\E[r^2(\xx)]$ that is independent of $\delta(\xx)$---to minimizing the fully \emph{computable} objective
\begin{equation}
    L(\delta) = \E_{\xx}\!\Big[\delta^2(\xx) + 2\,\inner{\grad \delta(\xx)}{\vv(\xx)}\Big].
\label{eq:stein-loss}
\end{equation}
 $L(\delta)$ uses only samples $\xx\!\sim\! p$ and evaluations of the vector field $\vv(\xx)$. Its minimizer is $\delta^\star(\xx)=\argmin_{\delta}L(\delta)\approx r(\xx)$, so we can approximate the divergence using the learned $\delta^*(\xx)$:
\begin{equation}
    \inner{\nabla}{\vv(\xx)} \approx \underbrace{b(\xx)+\delta^\star(\xx)}_{\text{baseline + residual}}.
    \label{eq:baseline+residual}
\end{equation}
We will now apply this framework to diffusion models (\S\ref{sec:stad-diffusion}), and flow matching/rectified flow models (\S\ref{sec:stad-flow-matching}).

\subsection{StAD for diffusion}
\label{sec:stad-diffusion}
For a diffusion model indexed by time $t\in[\varepsilon,T]$, we can first define a set of instantaneous quantities,
\begin{align}
    &\score_t(\xx_t) = \grad\log p_t(\xx_t),\\
    &b_t(\xx_t) = -\inner{\vv_t(\xx_t)}{\score_t(\xx_t)},
\end{align}
where, $p_t(\xx_t)$ is the marginal density at time $t$, $\vv_t(\xx_t)$ is the PF-ODE, $\score_t(\xx_t)$ is the pre-trained score network, and $b(\xx_t)$ is the Stein baseline. We learn a \emph{time-dependent} scalar head $\delta(\xx_t;t)$ to approximate the divergence residual, by averaging the loss in \eqref{eq:stein-loss} across times. We can define an instantaneous loss,
\begin{equation}
    L(\delta;t) =\E_{\xx_\varepsilon}\E_{\xx_{t}|\xx_{\varepsilon}}\!\Big[\delta^2(\xx_t;t) + 2\,\inner{\grad \delta(\xx_t;t)}{\vv_t(\xx_t)}\Big],
\end{equation}
as an average over the conditional distribution $p_{t}(\xx_t|\xx_\varepsilon)$ (the noising process, which can be easily sampled in closed form; see e.g. \citealp{song2021sde}), and the target $p(\xx_\varepsilon)$. Our total loss will be the time average of this,
\begin{equation}
    L(\delta) = \E_{t}\big[L(\delta;t)\big]
    \label{eq:total-loss}
\end{equation}
with $p(t)=U(\varepsilon,T)$ uniform between $\varepsilon$ and $T$ (although see Appendix \ref{sec:time-sampling}). We minimize this loss to find the optimum: $\delta^\star(\xx;t) = \argmin_{\delta} \big[L\big(\delta(\xx_t;t)\big)\big]$.

At test time, we simply replace $\inner{\nabla}{ \vv_t(\xx_t)}$ inside \eqref{eq:lik} by the learned surrogate $b_t(\xx_t)+\delta^\star(\xx_t;t)$ and integrate a \emph{scalar} log-ODE alongside the state ODE from $t=\varepsilon$ to $t=T$,
\begin{equation}
    \frac{\dd\xx_t}{\dd t} = \vv_t(\xx_t), \quad
    \frac{\dd \ell_t}{\dd t} = b_t(\xx_t)+\delta^\star(\xx_t;t), \label{eq:log-ode}
\end{equation}
with initial values at $t=\varepsilon$ of $\xx_\varepsilon$ (the known sample whose log likelihood we wish to evaluate) and $\ell_\varepsilon=0$. Solving the two ODEs for $\xx_T$ and $\ell_T$ gives our log likelihood:
\begin{equation}
    \log p_\varepsilon(\xx_\varepsilon) \approx \log p_T(\xx_T) + \ell_T
    \label{eq:log-like}
\end{equation}
For a conditional setting, we can follow the same approach as above, but with a context/conditioning variable, $z$, introduced. In such a setting, $\vv_t$, $\score_t$, $b$, and $\delta$ will all be modified to take $z$ as an input, $p(\xx_\varepsilon)$ will become $p(\xx_\varepsilon|z)$, and the loss will be additionally averaged over $z$. In both cases, the runtime cost is just forward passes through the velocity field $\vv_t$, the teacher score $\score_t$ for $b_t$, and the small head $\delta$; no Jacobian traces are needed.

\subsection{StAD for flow-matching and rectified flows}
\label{sec:stad-flow-matching}
The obvious challenges to extending \textit{StAD} to other generative models are the availability of a score for the baseline, and ensuring the vector field is of Stein class. Although flow-matching and rectified flow models \citep{lipman2023flowmatchinggenerativemodeling, liu2023flow} were developed as a different class of models to diffusion, it has been shown recently \citep{kingma2023diffusionfm, zhou2025scoredistillationflowmatching, albergo2025stochasticinterpolantsunifyingframework} that flow-matching models are equivalent to diffusion under Gaussian conditions (see also the work of \citealp{karras2022edm, karras2024edm}). 

\citet{zhou2025scoredistillationflowmatching} define the Gaussian score identity as
\begin{equation}
    \label{eq:gaussian-score}
    \textbf{s}_t(\xx_t) = \grad_{\xx_t} \log p_t(\xx_t) = -\frac{\xx_t - \alpha_t\E[\xx_\epsilon | \xx_t]}{\sigma_t^2} ,
\end{equation}
where $\alpha_t$ and $\sigma_t$ define the noising schedule, with
\begin{equation}
    \label{eq:noisingschedule}
    p(\xx_t|\xx_\varepsilon) = N(\alpha_t\xx_\epsilon, \sigma^2_t\mathbf{I}).
\end{equation}
defining the noising process. The most common configuration is to set $\alpha_t=1-t$ and $\sigma_t=t$.
The score in \eqref{eq:gaussian-score} is equivalent to Tweedie's formula \citep[e.g.][]{robbins1992empirical, efron2011tweedie}. The conditional flow-matching objective \citep{lipman2023flowmatchinggenerativemodeling},
\begin{equation}
    \label{eq:flow-matching-obj}
    \mathcal{L}(\theta) = \E_{t}\,\E_{\xx_\varepsilon}\E_{\xx_t|\xx_\varepsilon}\big[||\vv_t(\xx_t;\theta) - \mathbf{u}_t(\xx_t|\xx_\epsilon)||^2\big],
\end{equation}
where $\mathbf{u}(\xx_t|\xx_\varepsilon)$ is the conditional probability path defined by the noising process and $p(t)= U(\varepsilon,T)$, is used to learn a vector field $\vv_t(\xx_t;\theta)$ with parameters $\theta$. For this objective, and under the linear interpolation path where $\alpha_t + \sigma_t = 1$, \citet{zhou2025scoredistillationflowmatching} show that \eqref{eq:gaussian-score} is equivalent to,
\begin{equation}
    \score(\xx_t) = -\frac{\xx_t + \alpha_t\vv_t(\xx_t)}{\sigma_t(\alpha_t + \sigma_t)}.
    \label{eq: flow-initial-score}
\end{equation}

Plugging this into the baseline \eqref{eq: baseline} gives us a new Stein baseline for this class of model:
\begin{equation}
        b(\xx_t) = -\bigg\langle\vv_t(\xx_t),-\frac{\xx_t + \alpha_t\vv_t(\xx_t)}{\sigma_t(\alpha_t+\sigma_t)}\bigg\rangle. 
    \label{eq:flow-baseline}
\end{equation}
Similarly, for the target residual $r(\xx_t)$, we replace the score in \eqref{eq:residual} with \eqref{eq: flow-initial-score} to obtain,
\begin{equation}
        r(\xx_t) = \inner{\nabla}{\vv_t(\xx_t)} - \frac{\inner{\vv_t(\xx_t)}{\xx_t}+ \alpha_t||\vv_t(\xx_t)||^2}{\sigma_t(\alpha_t + \sigma_t)}. 
    \label{eq:flow-residual}
\end{equation}
After defining our baseline and residual, the rest of the \textit{StAD} objective remains the same as for diffusion models (\S\ref{sec:stad-diffusion}). Assuming our learned vector field is in Stein class, the Stein identity holds and gives us the same objective for $\delta$:
\begin{equation}
    L(\delta;t) = \E_{\xx_\varepsilon}\E_{\xx_t|\xx_\varepsilon}\!\Big[\delta^2(\xx_t;t) + 2\,\inner{\grad \delta(\xx_t;t)}{\vv_t(\xx_t)}\Big],
\end{equation}
We optimize our scalar head $\delta(\xx_t;t)$, by minimizing the time average of $L(\delta;t)$, 
and approximate the target divergence by $\inner{\nabla}{\vv_t(\xx_t)} \approx b(\xx_t)+\delta^\star(\xx_t;t)$. A very similar approach can be taken for TrigFlow \citep{lu2025simplifying}---a class of continuous-time consistency model---which we demonstrate in Appendix \ref{sec:stad-extensions}. 

\subsection{Regularising $\delta$ to ensure validity of the StAD loss}
\label{sec:regularisation}
We have derived \textit{StAD} on the conditions that the models we are distilling are Gaussian bridges, and that the learned vector field is in Stein class. Appendix \ref{sec:stein-class} examines the conditions under which $\vv_t(\xx_t)$ will satisfy Stein boundary conditions \citep{liu2016ksd, liu2016svgd, gorham2017ksd, gorham2019} for diffusion and flow-matching models. 

In our derivation of the \textit{StAD} loss, we have assumed that the boundary flux in \eqref{eq:vanishingmarginal} must vanish as $R \rightarrow \infty$.
We can regularise $\delta$ such that this occurs by construction. We define a regularised $\hat{\delta}(\xx) \coloneqq \kappa_R(\xx)\delta(\xx)$, with
\begin{equation}
\label{eq:cutoff}
    \kappa_R = \begin{cases} 
    1; &\hphantom{R <}\;\lvert \lvert \xx \rvert \rvert \le R \\ 
    \frac{1}{2}+\frac{1}{2}\cos\big(\frac{\pi}{R}\lvert \lvert \xx \rvert \rvert -\pi\big);  &R < \lvert \lvert \xx \rvert \rvert  <2R \\
    0; &\hphantom{R <}\;\lvert\lvert\xx\rvert\rvert \ge 2R
    \end{cases}
\end{equation}
Between $2R$ and $R$, $\kappa_R$ is a smooth transition between 1 and 0, using a cosine function. We set $R$ such that $\mathbb{P}_{\xx\sim p}(\lvert\lvert\xx\rvert\rvert > R)$ is negligible. This can be achieved by setting $R$ to a high quantile of $\lvert\lvert\xx\rvert\rvert$. This creates a compact support for $\hat{\delta}$ and forces it to decay quickly. The only caveat here, is that if some probability mass lives outside of $2R$, it will be neglected. Now, that $\hat{\delta}(\xx)$ has compact support, the boundary-flux term in \eqref{eq:vanishingmarginal}, $p(\xx)\,\delta(\xx)\,\vv(\xx)$ vanishes regardless of the properties of $\vv(\xx)$ or $p(\xx)$. We can define the ball $B$ on $2R$, and apply the divergence theorem as in \eqref{eq:vanishingmarginal}, seeing that the boundary flux vanishes because, $p(\xx)\hat{\delta}(\xx)\vv(\xx) = 0 \quad \forall \lvert\lvert\xx\rvert\rvert \ge 2R$.

Since, $\hat\delta(\xx) = \kappa_R\delta(\xx)$, its derivative with respect to $\xx$ is
\begin{equation}
    \grad \hat\delta(\xx) = \kappa_R(\xx)\grad\delta(\xx) + \delta(\xx)\grad\kappa_R(\xx).
\end{equation}
This gives us a new regularised \textit{StAD} objective:
\begin{multline}
\label{eq:regularisedobjective}
        L(\delta) = \E_{\xx}\!\Big[\kappa^2_R(\xx)\delta^2(\xx) + 2\kappa_R(\xx)\inner{\grad \delta(\xx)}{\vv(\xx)} \\+ 2\delta(\xx)\inner{\grad \kappa_R(\xx)}{\vv(\xx)}\Big].
\end{multline}
Since we define the transition in \eqref{eq:cutoff} to be cosine, $\grad \kappa_R$ when $R < \lvert\lvert\xx\rvert \rvert < 2R$  can be computed as,
\begin{equation}
\grad \kappa_R(\xx) = - \frac{\pi}{2R}\frac{\xx}{\lvert\lvert\xx\rvert \rvert} \sin \Big (\frac{\pi}{R}\lvert\lvert\xx\rvert \rvert - \pi\Big ),
\end{equation}
and is zero otherwise.

In practice we include a further $L^2$ penalty in the loss by adding $\E_\xx\big[l||\nabla\hat{\delta}(\xx)||^2\big]$ to \eqref{eq:regularisedobjective}. Here $l$ can be set to regularise the gradient of $\hat{\delta}(\xx)$ and ensure smoothness. We find in our experiments in \S\ref{sec:experiments} that the regularised objective gives more consistent outcomes, since the boundary condition assumed in \eqref{eq:vanishingmarginal} holds by construction. 

\section{Experiments and results}
\label{sec:experiments}
We test \textit{StAD} on a variety of density estimation tasks using diffusion models, to evaluate its efficacy on low- to high-dimensional problems. We report the variance, mean-absolute error, and wall time for log-likelihood estimation compared to the exact trace, Hutchinson, Hutch++, and XTrace algorithms when predicting astrophysical fluxes and flux errors for the Cosmic Evolution Survey \citep[\texttt{COSMOS2020};][]{scoville2007cosmos, weaver2022cosmos} in \S\ref{sec:cosmos}. In \S\ref{sec:image}, we report negative log-likelihood (NLL) estimates along with variance and relative speedups on the \texttt{CIFAR-10} \citep{krizhevsky2009cifar} and \texttt{ImageNet-32x32} \citep{chrabaszcz2017imagenet} datasets to show the scalability of our method. Full practical details of the experiments are given in Appendix \ref{sec: detailedexperiments}

\subsection{Astrophysical fluxes}
\label{sec:cosmos}
\begin{figure}
  \begin{center}
\centerline{\includegraphics[width=\columnwidth]{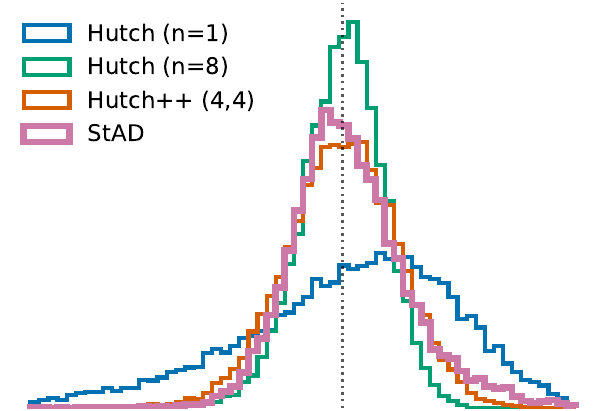}}
    \caption{ Divergence estimation for a VP-SDE diffusion model, trained to predict astrophysical fluxes and flux errors (26 dimensional conditional density). Histograms show log likelihood residuals (i.e.\ $\text{exact}-\text{estimated}$ log likelihood) for Hutchinson with 1 and 8 probes, rank-4 Hutch++ with 4 probes, and \textit{StAD}. Dotted line corresponds to zero residual.}
    \label{img:astroresiduals}
  \end{center}
\end{figure}

\begin{figure}
  \begin{center}
\centerline{\includegraphics[width=\columnwidth]{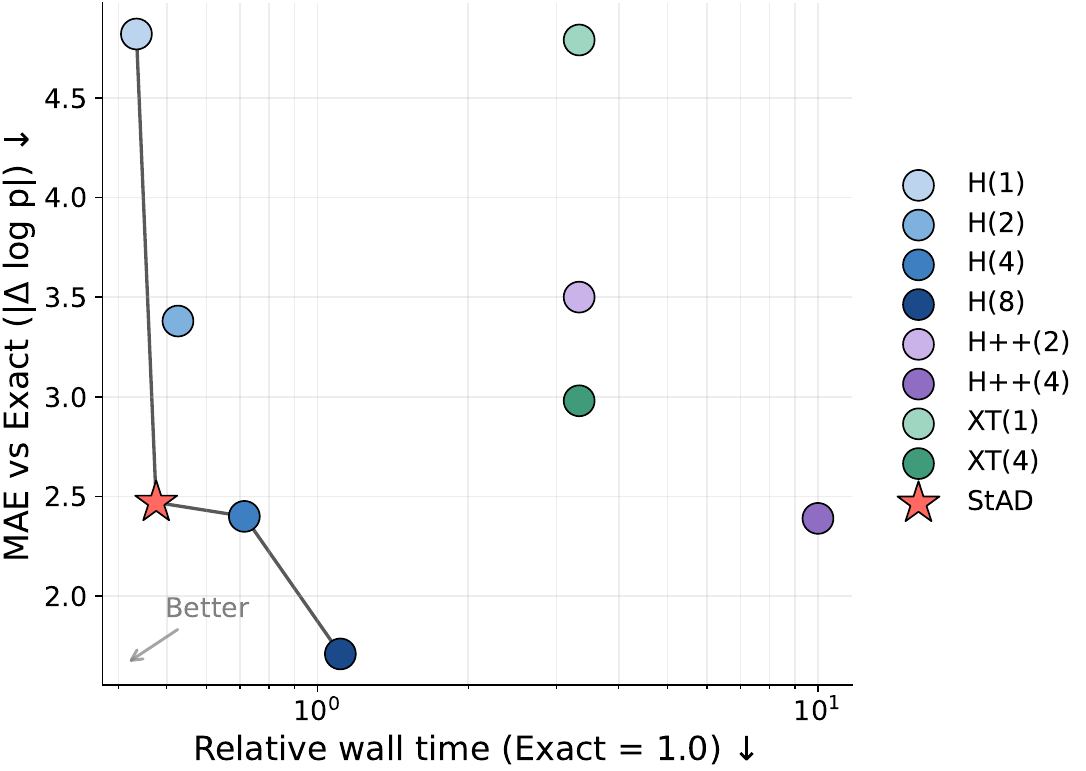}}
    \caption{ Pareto optimal divergence estimation with PF-ODEs. These are estimates for a VPSDE diffusion model predicting a 26 dimensional conditional density over astrophysical fluxes and flux errors. The figures plotted here are the same as in \cref{tab:cosmos}.
    }
    \label{img:astropareto}
  \end{center}
\end{figure}

Simulation based inference and forward modelling for astronomical surveys relies on having a realistic generative model for the uncertainties on flux measurements \citep[see e.g.][]{alsing2023forwardmodel, Alsing_2024, crenshaw2024forwardmodel, Thorp_2025}, which is typically learned from data. As a test of \textit{StAD}, we train a model for flux uncertainty conditional on magnitude (logarithmically scaled flux) using the \texttt{COSMOS2020} dataset \citep{weaver2022cosmos}. This is a $26\times26$ conditional density estimation problem, with the model learning the distribution of 26 flux uncertainties conditional on 26 magnitudes (coming from 26 wavelength filters used in the COSMOS survey). The training data comprises $\sim420,000$ galaxies from \texttt{COSMOS2020}.

We train a variance preserving-SDE (VP-SDE) diffusion model that learns the conditional density \citep{Thorp_2025}; and distill it with \textit{StAD}. The original \textit{teacher} VP-SDE is trained via denoising score matching, and uses a $256 \times 5$ multi-layered perceptron (MLP) as the score model. We use a $512 \times3$ MLP to distill the PF-ODE and learn a divergence estimator. We use the distilled model to estimate conditional probabilities for a set of test samples from \texttt{COSMOS2020}. We apply the Hutchinson, Hutch++, and XTrace estimators with different numbers of probes to the original VP-SDE, and compare all estimators to the exact trace, with results shown in \cref{tab:cosmos}, \cref{img:astroresiduals}, and \cref{img:astropareto}. 

We find that \textit{StAD} outperforms the stochastic estimators consistently in wall-time, with the exception of single-probe Hutchinson. \textit{StAD} shows remarkably low mean-absolute-error (MAE) relative to the exact log likelihoods, as shown in \cref{tab:cosmos}. The only estimators that have a lower MAE and variance, are not much faster than the exact trace: Hutchinson with 4 or 8 probes and rank-4 Hutch++ or XTrace with 4 probes. The mean and standard deviation reports in \cref{tab:cosmos} show that \textit{StAD} is slightly biased compared to other trace estimators. 

For Hutch++, we implement the \texttt{FFJORD++} algorithm as shown in \citet{liu2025trace}, where we cache the $Q$ matrix for the $QR$ decomposition and refresh it every $6$ iterations. However, we still find that computing $QR$ decomposition dominates the cost and thus we find the Hutch++ to be even relatively slow in our tests. We plot the distribution of residuals for a subset of the estimators in \cref{img:astroresiduals}, noticing that the \textit{StAD} residuals have a slight tail towards overestimates of the log likelihood (but less skew and variance than the Hutchinson estimator). The size of the distillation network relative to the teacher directly affects the inference time for \textit{StAD}; since the teacher network here was relatively compact, the relative speedups are modest. However, even though \textit{StAD} has the same number of NFEs as the exact trace (seen in \cref{tab:cosmos}), it is still faster since we do not compute the Jacobian or its trace.

\begin{table}[t]
    \centering
    \caption{Comparing the Hutchinson, Hutch++ and \textit{StAD} log-likelihood estimates, to the exact trace for a diffusion model that predicts astrophysical flux uncertainties (\S\ref{sec:cosmos}). Methods: H($n$) $=$ Hutchinson with $n$ probes; H++($n$) $=$ rank-$n$ Hutch++ with $n$ probes; XT($n$) $=$ rank-$n$ XTrace with $n$ probes; \textit{StAD} $=$ this work. Mean, standard deviation, MAE, and speedup are all measured relative to exact trace. NFEs are measured relative to the exact trace: $\text{rNFEs}=\text{NFEs [approx]}/\text{NFEs [exact]}$.}
    \label{tab:cosmos}
    \small
    \begin{tabular}{lrccc}
    \toprule
    \textbf{method} &
    \textbf{mean $\pm$ std} &
    \textbf{speedup ($\times$)} &
    \textbf{MAE} &
    \textbf{rNFEs} \\
    \midrule
    H(1) & $0.06 \pm 6.12$ & $2.3$  & $4.82$ & $0.85$ \\
    H(2) & $0.03 \pm 4.37$ & $1.9$  & $3.38$ & $0.83$\\ 
    H(4) & $0.00 \pm 3.08$ & $1.4$  & $2.40$ & $0.82$ \\
    H(8) & $0.00 \pm 2.14$ & $0.9$  & $1.71$ & $0.83$\\
    H++(2) & $-0.08 \pm 4.46$ & $0.3$  & $3.50$ & $0.82$ \\
    H++(4) & $0.02 \pm 2.98$ & $0.1$  & $2.39$ & $0.82$\\
    XT(1) & $-0.00 \pm 6.15$ & $0.3$ & $4.79$ & $0.82$ \\
    XT(4) & $-0.00 \pm 3.74$ & $0.3$ & $2.98$ & $0.82$ \\
    \textit{StAD} & $0.46 \pm 3.41$ & $2.1$ & $2.47$ & $1.00$\\
    \bottomrule
    \end{tabular}
\end{table}

\subsection{CIFAR-10 and ImageNet-32$\times$32}
\label{sec:image}
We train a diffusion model with VP-SDE and U-Net architecture \citep{ronneberger2015unet} on \texttt{CIFAR-10} and \texttt{ImageNet-32x32}. To distill these models, we use a compact convolutional neural network (CNN; \citealp{lecun1990, lecun89}) with a linear time-embedding and Feature-wise Linear Modulation (FiLM; \citealp{perez2017filmvisualreasoninggeneral}) conditioning mechanism to inject time information into the spatial layers. We describe the architecture in more detail in Appendix \ref{sec: detailedexperiments}. 

We compute negative log-likelihood (NLL) estimates from these models under matched ODE solver settings, expressing the results for each image in bits per dimension,
\begin{equation}
    bpd(\xx_\varepsilon) = \frac{-\log p(\xx_\varepsilon)}{D\log2} +\text{offset}.
\end{equation}
Here we have an offset of $7$, since the images are treated as continuous in $[-1,1]$ in the ODE, but in reality have an 8-bit discrete structure \citep{theis2016noteevaluationgenerativemodels}. Hence we report continuous NLL in $bpd$ with the discretization offset, and add uniform dequantisation noise to the images. We evaluate NLLs using \textit{StAD} and Hutchinson for a test set of 2048 images, computing the mean and standard deviation of NLL estimates across the sample. The results are summarized in \cref{tab:CIFAR} and \cref{tab:ImageNet}.

\begin{figure}
  \begin{center}
\centerline{\includegraphics[width=\linewidth]{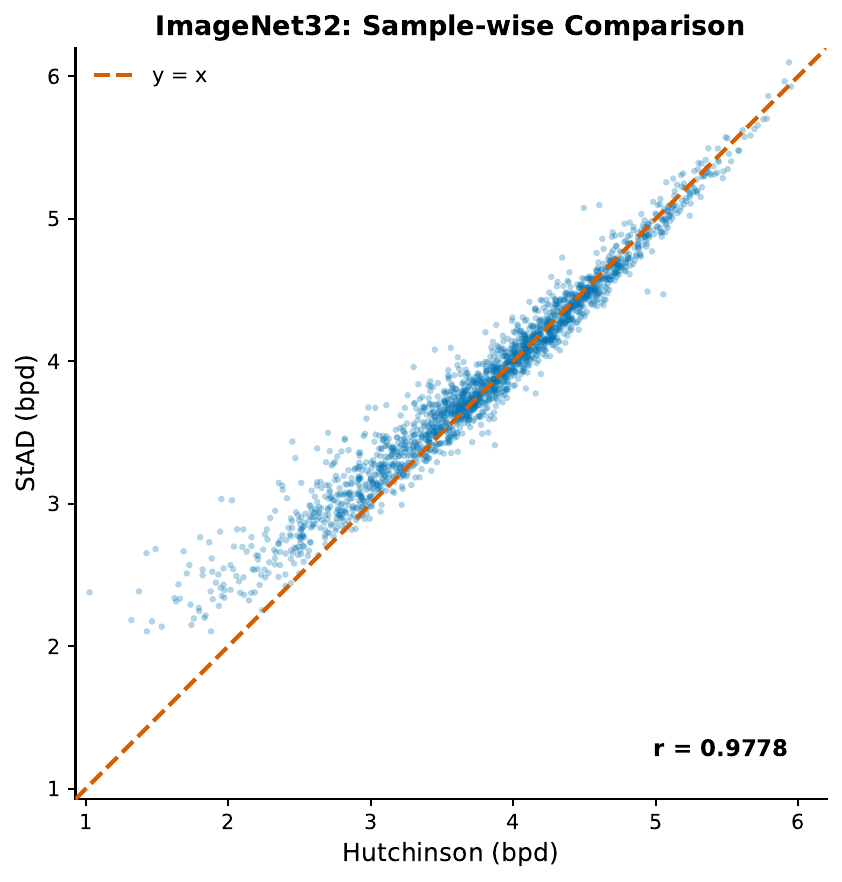}}
    \caption{Sample-wise correlation test for NLL estimates from \textit{StAD} vs.\ 2-probe Hutchinson, for 2048 images taken from \texttt{ImageNet32x32}. Each blue point corresponds to one image.
    }
    \label{img:ImageNet_scatter}
  \end{center}
\end{figure}

We show that \textit{StAD} remains competitive with the Hutchinson trace estimator, with consistent estimation of the mean NLL, whilst significantly reducing compute-time and complexity. For \texttt{CIFAR-10}, we observe that the difference in the mean NLL estimates is $\sim0.02~bpd$. Similarly, for \texttt{ImageNet-32x32}, we observe a difference in the mean NLL of $\sim0.07~bpd$. The NLL estimates from \textit{StAD} show less variance when averaged across the test set. Additionally, in \cref{img:ImageNet_scatter} we perform a correlation test on a set of $2048$ images, to compare the NLL estimates from Hutchinson with 2 probes against \textit{StAD} on an image-by-image basis. We find for \texttt{ImageNet-32x32} that the correlation coefficient is as high as $\sim 0.98$.

\begin{figure*}[ht]
  \begin{center}
\centerline{\includegraphics[width=\linewidth]{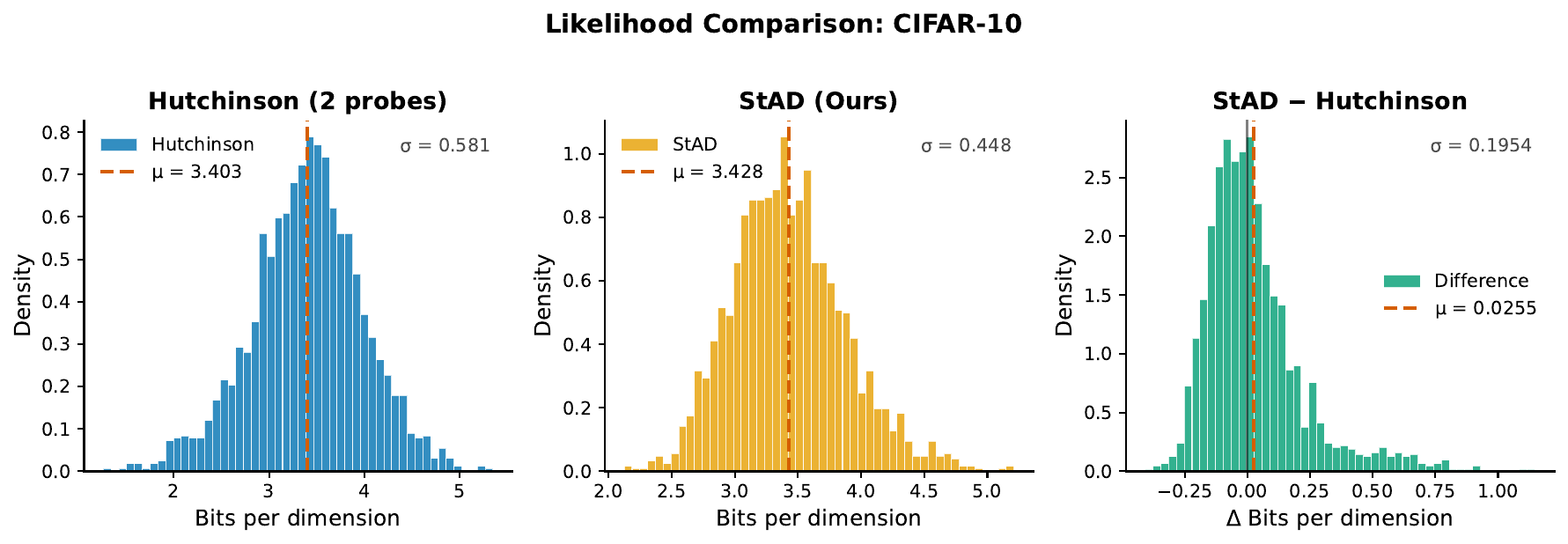}}
    \caption{
    Sample-wise NLL estimates for \texttt{CIFAR-10}. \textit{Left}: Using the Hutchinson algorithm with 2 probes. 
    \textit{Centre}: Using \textit{StAD}. \textit{Right}: Image-by-image difference between \textit{StAD} and Hutchinson with 2 probes: $\text{NLL}[\textit{StAD}] - \text{NLL}[\text{H(2)}]$.}
    \label{img:CIFAR_bpd}
  \end{center}
\end{figure*}

As seen in \cref{img:CIFAR_bpd}, for \texttt{CIFAR-10} the difference in NLL estimates between the two methods is generally small, albeit with some image-to-image scatter. We compute the amount of times the ODE drift term is evaluated (giving the number of NFEs) when integrating the likelihood; \cref{tab:CIFAR} and \cref{tab:ImageNet} show that \textit{StAD} has a factor of $\sim4\times$ fewer NFEs than the single-probe Hutchinson estimator. Moreover, bypassing the Jacobian vector product required by Hutchinson leads to an additional speedup of $\sim2\times$ on top of the $\sim4\times$ reduction in NFEs. Further results for these datasets are included in Appendix \ref{sec:image-results-extra} and \ref{sec:horse}.

\begin{table}[t]
    \centering
    \caption{Comparing the Hutchinson and \textit{StAD} NLL estimates for a VP-SDE diffusion model trained on \texttt{CIFAR-10}. NLLs are the mean and standard deviation across a test set of 2048 images. Speedup is measured relative to single-probe Hutchinson---H(1). NFEs are averaged over the total number per batch of 512 images.}
    \label{tab:CIFAR}
    \small
    \begin{tabular}{llccc}
    \toprule
    \textbf{method} &
    \textbf{NLL ($bpd$)} &
    \textbf{speedup ($\times$)} &
    \textbf{NFEs} \\
    \midrule
    H(1)       & $3.403 \pm 0.583$ & $1.00$ & $980$ \\
    H(2)       & $3.402 \pm 0.580$ & $0.65$ & $970$ \\%
    \textit{StAD}        & $3.428 \pm 0.44$ & $7.24$ & $250$ \\
    \bottomrule
    \end{tabular}
\end{table}
\begin{table}[t]
    \centering
    \caption{Comparing the Hutchinson and \textit{StAD} NLL estimates for a diffusion model trained on \texttt{ImageNet-32x32}. Column definitions are as in \cref{tab:CIFAR}.}
    \label{tab:ImageNet}
    \small
    \begin{tabular}{llccc}
    \toprule
    \textbf{method} &
    \textbf{NLL ($bpd$)} &
    \textbf{speedup ($\times$)} &
    \textbf{NFEs} \\
    \midrule
    H(1)       & $3.783 \pm 0.787$ & $1.00$ & $990$ \\
    H(2)       & $3.782 \pm 0.786$ & $0.65$ & $981$ \\%
    \textit{StAD}        & $3.856 \pm 0.673$ & $7.56$ & $240$ \\
    \bottomrule
    \end{tabular}
\end{table}

\section{Discussion and conclusion}
\label{sec:conclusion}
We present the \textit{Stein Amortized Divergence} or \textit{StAD} as a distillation technique to achieve pareto-optimal likelihood evaluations (\cref{img:astropareto}) for continuous time flows. We show that our method can be generalised to a range of different generative paradigms (\S\ref{sec:methodology} and Appendix \ref{sec:stad-extensions}), and demonstrate the efficacy of our method on diffusion workflows. \textit{StAD} is a deterministic estimator of the divergence of the learned vector field, and thus provides smoother ODE trajectories, yielding faster likelihoods in addition to not requiring the Jacobian or its trace during inference. This is in contrast to scalable but stochastic estimators like the Hutchinson, Hutch++, and XTrace that still require Jacobian--vector products. 

In workflows where frequent evaluations of the likelihood are needed, \textit{StAD} is a promising alternative to traditional trace estimators. \textit{StAD} is however not favourable when rare or sparse evaluations of the likelihood are needed, since it needs additional compute to distill the model in the first place. We find that the speedups achieved using \textit{StAD} (compared to a stochastic estimator) are relative to the discrepancy between the student and teacher networks---a single JVP using the teacher network is of similar computational complexity to the forward pass through the student, assuming the architectural complexity is similar. If the student network is considerably more compact than the teacher, the potential gains from \textit{StAD} distillation are much larger. This is seen in the discrepancy in speedups between \cref{tab:cosmos} and \cref{tab:CIFAR} or \cref{tab:ImageNet} along with the difference in NFEs.

Future applications and tests of \textit{StAD} are in: applying it to Boltzmann Generators \citep{noé2019boltzmanngeneratorssampling, rehman2025falconfewstepaccuratelikelihoods}, where rapid and accurate likelihoods are necessary for importance weighting; aiding other distillation procedures and constructing flow-maps such as in \citet{ai2025distillation}; improving flow and diffusion based modelling \citep{huang2025improving}; guidance of diffusion models \citep{karczewski2025diffusion}; 
black-box Bayesian optimisation as show in \citet{yun2025posterior}; and out-of-distribution detection frameworks as shown in \citet{raonic2026towards}, to name a few.






\section*{Acknowledgements}
We thank Anik Halder and Justin Alsing for useful comments regarding the project and some of the source code. This work has been supported by funding from the European Research Council (ERC) under the European Union's Horizon 2020 research and innovation programmes (grant agreement no.\ 101018897 CosmicExplorer and no.\ 818085 GMGalaxies), and the research project grant ``Understanding the Dynamic Universe'' funded by the Knut and Alice Wallenberg Foundation under Dnr KAW 2018.0067. This research utilized the Sunrise HPC facility supported by the Technical Division at the Department of Physics, Stockholm University.

\section*{Impact Statement}
This paper presents work whose goal is to advance the field of Machine Learning. There are many potential societal consequences of our work, none of which we feel must be specifically highlighted here.

\section*{LLM Usage}
We used LLMs to assist with code generation, verification, and as an initial research aid. The authors take full responsibility for the materials provided with this body of work.


\makeatletter\g@addto@macro{\UrlBreaks}{\do\/\do\-}\makeatother

\bibliography{example_paper}
\bibliographystyle{icml2026}

\appendix
\section{Stein class conditions}
\label{sec:stein-class}
The boundary condition for the Stein class \citep{liu2016svgd, gorham2017ksd, barp2019msd} sets the requirements on $\vv_t(\xx_t)$ for Stein's identity \eqref{eq:stein-identity} to hold when applying the Langevin--Stein operator \eqref{eq:langevin-stein} to $\vv_t(\xx_t)$. The requirement can be expressed for $\xx_t\in\R^D$ \citep[see e.g.][]{liu2016ksd} as a vanishing boundary flux,
\begin{equation}
    \lim_{R\to\infty}\oint_{\partial B_R} \inner{p_t(\xx_t)\,\vv_t(\xx_t)}{\hat{\mathbf{n}}}\,\dd S\to0,
    \label{eq:surface-integral}
\end{equation}
where $\hat{\mathbf{n}}=\xx_t/||\xx_t||\in\R^D$ is the outward unit normal to the ball $B_R$ with $R=||\xx_t||$, analogously to \eqref{eq:vanishingmarginal}.

\subsection{Diffusion models}
For diffusion models of the class considered by \citet{song2021sde}, $\vv_t(\xx_t)=\mathbf{f}(\xx_t,t) - \frac{1}{2}g^2(t)\,\grad\log p_t(\xx_t)$, which gives
\begin{multline}
    p_t(\xx_t)\,\vv_t(\xx_t) = p_t(\xx_t)\mathbf{f}(\xx_t,t) -\frac{1}{2}g^2(t)\grad p_t(\xx_t)\\
    =\E_{\xx_\varepsilon}\Big[p_t(\xx_t
    |\xx_\varepsilon)\mathbf{f}(\xx_t,t) -\frac{1}{2}g^2(t)\grad p_t(\xx_t|\xx_\varepsilon)\Big],
    \label{eq:diffusion-boundary-integrand}
\end{multline}
at the boundary. For the (sub-)VP-SDE, we have $\mathbf{f}(\xx_t,t)=-\frac{1}{2}\beta(t)\xx_t$, giving $\inner{\mathbf{f}(\xx_t,t)}{\hat{\mathbf{n}}}=-\frac{1}{2}\beta(t)$, whilst for the variance exploding (VE)-SDE, we have $\mathbf{f}(\xx_t,t)=0$. The transition probabilities, $p(\xx_t|\xx_\varepsilon)$ are Gaussian, with general form $N\big(\nu(t)\xx_\varepsilon, \eta^2(t)\mathbf{I}\big)$, where $\nu(t)$ and $\eta(t)$ are scalar-valued functions of $t$. This gives
\begin{equation}
    \grad p_t(\xx_t|\xx_\varepsilon)=-\frac{p(\xx_t|\xx_\varepsilon)}{\eta^2(t)}\,\big(\xx_t-\nu(t)\xx_\varepsilon\big),
\end{equation}
and thus
\begin{equation}
    \inner{\grad p_t(\xx_t|\xx_\varepsilon)}{\hat{\mathbf{n}}} = -\frac{p(\xx_t|\xx_\varepsilon)}{\eta^2(t)}\big(1-\nu(t)\,\inner{\xx_\varepsilon}{\hat{\mathbf{n}}}\big).
\end{equation}
By Cauchy--Schwarz (see e.g.\ \citealp{vershynin2018hdp}), we can bound the inner product between $\xx_\varepsilon$ and the outward norm, $|\inner{\xx_\varepsilon}{\hat{\mathbf{n}}}|\leq||\xx_\varepsilon||$. The terms in the integrand of \eqref{eq:surface-integral} consequently take the form $\E_{\xx_\varepsilon}\big[p_t(\xx_t|\xx_\varepsilon)\big]$, or can be upper-bounded by $\E_{\xx_\varepsilon}\big[p_t(\xx_t|\xx_\varepsilon)\,||\xx_\varepsilon||\big]$, up to $t$-dependent constants. Transforming to hyper-spherical coordinates \citep[e.g.][]{song2020score}, we have
\begin{equation}
    \begin{split}
        p_t(\xx_t|\xx_\varepsilon) &\propto R^{D-1} \exp\bigg(\frac{-||\xx_t-\nu(t)\xx_\varepsilon||^2}{2\eta^2(t)}\bigg)\\
        & \leq R^{D-1} \exp\Bigg(\frac{-\big(R-\nu(t)||\xx_\varepsilon||\big)^2}{2\eta^2(t)}\Bigg),
    \end{split}
    \label{eq:reverse-triangle}
\end{equation}
by the reverse triangle inequality for $||\xx_t-\nu(t)\xx_\varepsilon||$. If $||\xx_\varepsilon||<\infty~\forall\xx_\varepsilon$, $p(\xx_t|\xx_\varepsilon)$ and $||\xx_\varepsilon||\,p(\xx_t|\xx_\varepsilon)$ will vanish as $R\to\infty$, meaning the expectation over $\xx_\varepsilon$, and the full surface integral in \eqref{eq:surface-integral} will also vanish. We therefore expect that for VE-, VP-, and sub-VP-SDEs, and a well behaved target distribution, the vector field produced by a diffusion model should be Stein class $\forall t\geq\varepsilon$.

\subsection{Flow-matching models}
For flow-matching models (and related classes like TrigFlow that obey a similar noising schedule), we have
\begin{equation}
    p(\xx_t)\vv_t(\xx_t) = \E_{\xx_\varepsilon}\big[p(\xx_t|\xx_\varepsilon)\vv_t(\xx_t)\big],
\end{equation}
with the distribution of $\xx_t$ being Gaussian conditional on $\xx_\varepsilon$ per \eqref{eq:noisingschedule}. Using the spherical coordinate transform as in \citet{song2020score}, we can re-write the conditional distribution of $\xx_t$ as
\begin{equation}
    \begin{split}
        p(\xx_t|\xx_\varepsilon) &\propto R^{D-1} \exp\bigg(\frac{-||\xx_t-\alpha_t\xx_\varepsilon||^2}{2\sigma_t^2}\bigg)\\
        & \leq R^{D-1} \exp\Bigg(\frac{-\big(R-\alpha_t||\xx_\varepsilon||\big)^2}{2\sigma_t^2}\Bigg),
    \end{split}
\end{equation}
where we have used the reverse triangle inequality to lower-bound $||\xx_t-\alpha_t\xx_\varepsilon||^2$ as in \eqref{eq:reverse-triangle}. By Cauchy--Schwarz, we can bound $|\inner{\vv_t(\xx_t)}{\hat{\mathbf{n}}}|\leq||\vv_t(\xx_t)||$ as being less than or equal to the magnitude of the learned vector field. If $||\vv_t(\xx_t)||$ is bounded $\forall\xx_t$, or is limited to polynomial growth with $R$, the integrand in \eqref{eq:surface-integral} will vanish $\forall\xx_\varepsilon$ and the Stein class will be satisfied (c.f.\ also the conditions discussed in \S\ref{sec:stad-diffusion} and \S\ref{sec:regularisation}).

\section{Comparison of trace estimators}
\label{sec:hutch-comparison}
As a cross-check of our main results (\S\ref{sec:experiments}), we perform some simple experiments to test the scaling and performance of the stochastic trace estimators on randomly generated matrices. To test the algorithms on non-PSD matrices, we generate a random $D\times D$ matrix $\mathbf{A}$ with $A_{ij}\sim N(0,1)$ for $D=4,16,64,256$. To test performance on PSD matrices we take $A_{ij}\sim\text{Half-}N(0,1)$.

\begin{figure}
    \centering
    \includegraphics[width=\linewidth]{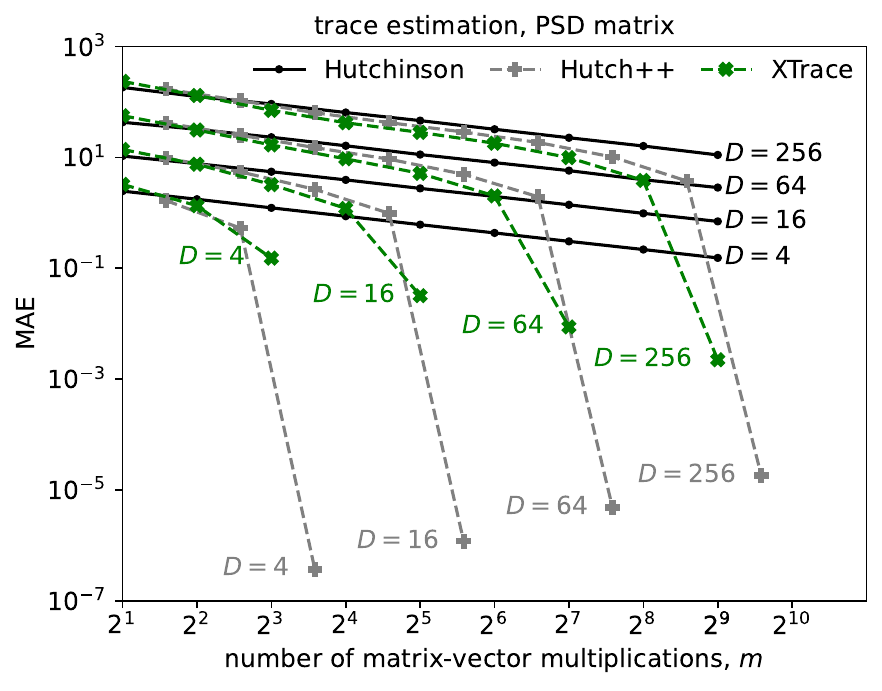}
    \includegraphics[width=\linewidth]{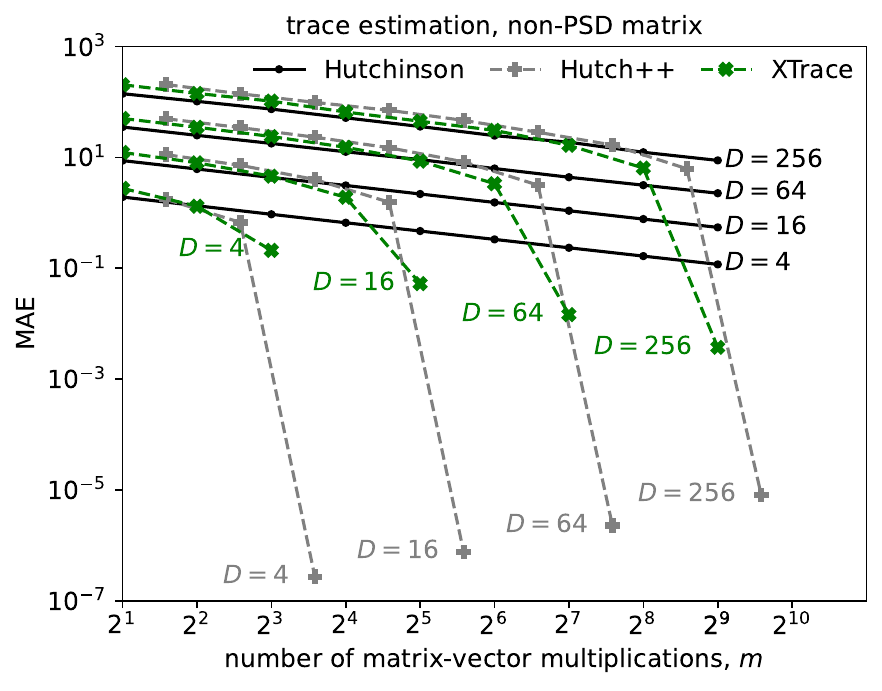}
    \caption{Stochastic trace estimation for a randomly generated $D\times D$ matrix $\mathbf{A}$. \textit{Top}: PSD $\mathbf{A}$ with $A_{ij}\sim\text{Half-}N(0,1)$. \textit{Bottom}: non-PSD $\mathbf{A}$ with $A_{ij}\sim N(0,1)$. MAE relative to exact trace is calculated by averaging over $B=65536, 16384, 4096, 1024$ matrices for $D=4,16,64,256$.}
    \label{fig:hutch-comparison}
\end{figure}

We compute the exact trace $\text{Tr}(\mathbf{A})=\sum_{i=1}^DA_{ii}$, and compare this to the trace computed using the stochastic estimators. We average over $B=65536, 16384, 4096, 1024$ random matrices for a given $D$ to calculate the MAE relative to the exact trace. For the Hutchinson estimator, we use $n\in[1,512]$ probe vectors, requiring $m=n$ matrix--vector multiplications for a given $n$. For Hutch++ and XTrace, we use $n\in[1,D]$ probe vectors with a rank-$n$ $QR$ decomposition, requiring respectively $m=3n$ or $m=2n$ matrix--vector multiplications.

Figure \ref{fig:hutch-comparison} shows the results of our experiment. For the PSD case, shown in the top panel, we see that Hutch++ and XTrace outperform Hutchinson for all $m\geq4$ irrespective of $D$, with XTrace being the most performant algorithm at fixed $m$. For the non-PSD case, the result is more dependent on $D$ and $m$. For $m\ll D$, the basic Hutchinson estimator tends to yield a lower MAE, as seen in many of our main experiments (\S\ref{sec:experiments}). We find that XTrace begins outperforming Hutchinson at lower $m$ than Hutch++ does, and consistently outperforms Hutch++ to a similar level as in the PSD case. We find that for $m\geq D$ (i.e.\ $n\geq D/2$) XTrace consistently yields a lower MAE than the basic Hutchinson estimator, with parity achieved at $m=D/2$ (i.e.\ $n=D/4$). For PSD and non-PSD matrices, we find that Hutch++ always converges to a very small MAE at $n=D$, aligning with the findings of \citet{meyer2021hutchoptimalstochastictrace} who prove that their estimator will always asymptotically beat Hutchinson.

\begin{table}
    \centering
    \caption{Training cost for \textit{StAD} compared to directly learning the divergence---H(1) or the residual---H(1)+B. Cache time covers the generation of targets (vector field evaluations for \textit{StAD}, divergence or residual estimates for H(1) or H(1)+B. Train time covers the training of the distillation network using the cached targets. Step time covers a single training epoch.}
    \label{tab:direct-divergence-time}
    \small
    \begin{tabular}{lcccc}
    \toprule
    \textbf{method} &
    \textbf{cache (}s\textbf{)} &
    \textbf{train (}s\textbf{)} &
    \textbf{total (}s\textbf{)} &
    \textbf{step (}s\textbf{)} 
    \\
    \midrule
    H(1)       & $172.9$ & $7770.5$ & $7943$ & $0.03$ \\
    H(1)+B       & $172.9$ & $7780.1$ & $7953$ & $0.03$ \\%
    \textit{StAD}        & $73.1$ & $	7780.6$ & $7854$ & $0.16$ \\
    \bottomrule
    \end{tabular}
\end{table}

\begin{table}
    \centering
    \caption{Comparing \textit{StAD} to directly learning the divergence---H(1) or the residual---H(1)+B. We use the same FiLM-CNN described in \cref{tab:CIFARdistillhp}. MAE is computed against log probabilities summed over 256 images with H(16). JVPs are per-target evaluation.}
    \label{tab:direct-divergence-mae}
    \small
    \begin{tabular}{lccccc}
    \toprule
    \textbf{method} &
    \textbf{JVP} &
    \textbf{steps} &
    \textbf{time (}s\textbf{)} &
    \textbf{MAE} &
    \textbf{mean $\pm$ std} 
    \\
    \midrule
    H(1)       & $1$ & $281091$ & $7943$ & $0.59$ & $0.58 \pm 0.33$\\
    H(1)+B       & $1$ & $279727$ & $7953$ & $0.35$  & $0.35 \pm 0.22$\\%
    \textit{StAD}        & $0$ & $50000$ & $7854$ & $0.24$ & $0.21 \pm 0.26$\\
    \bottomrule
    \end{tabular}
\end{table}

\section{Direct divergence regression}
\label{sec:direct_divergence}
\textit{StAD} avoids computing the Jacobian or its trace in the definition of its loss. However, directly regressing against divergence estimates from trace estimators (analogously to e.g.\ \citealp{ai2025distillation}) could be a viable alternative. To study how \textit{StAD} compares to directly learning the divergence or the residual (defined in Equation \ref{eq:residual}), we train a neural network to regress against divergence estimates from Hutchinson, denoted as H(1), or against residual estimates, denoted as H(1)+B. The computational cost of training the direct divergence regressors are given \cref{tab:direct-divergence-time}, with the log probability accuracy compared to \textit{StAD} in \cref{tab:direct-divergence-mae}.

In \cref{tab:direct-divergence-time}, we report the time taken to build up a cache of targets (either vector field evaluations for \textit{StAD}, or divergence/residual estimates for the direct regressors), the training time, the total compute time (training + cache), and the time taken per training epoch. We find that \textit{StAD} takes more seconds per epoch compared to directly learning the divergence. Hence, to maintain a fair comparison we match the wall-clock time across methods and find that for a comparable time budget \textit{StAD} achieves better MAE than learning the divergence or the residual directly. A natural objection is that averaging over $N>1$ Hutchinson probes would reduce target variance and achieve better results. However, this would scale cache cost linearly while reducing target noise only by $\sqrt{N}$, rapidly exceeding \textit{StAD}'s total cost without closing the accuracy gap.

\section{Time sampling}
\label{sec:time-sampling}
The time averaged objective functions for \textit{StAD} are computed by averaging over uniformly sampled times, $t\sim U(\varepsilon,T)$. This has the desirable property of weighting all time-steps equally in the total loss. However, our experiments showed that oversampling times closer to $\varepsilon$ (i.e.\ closer to the target distribution) produces empirically better results, with lower variance in the resulting trace estimator. This is likely due to the behaviour of the vector field becoming less complex as $t\to T$. 

To achieve the benefit of denser early-time sampling, without changing the meaning of the time-integrated loss, we use an importance re-weighting trick, similar to \citet{song2021likelihood}. We define a new proposal distribution, $q(t)\propto 1/t^2$ for $t\in[\varepsilon,T]$:
\begin{equation}
    q(t) = \begin{cases}
        t^{-2}(1/\varepsilon-1/T)^{-1}; &\text{if }\varepsilon\leq t\leq T\\
        0; &\text{otherwise}.
    \end{cases}
\end{equation}
We rewrite our old loss function as an expectation over $q(t)$, re-weighted by $w(t) = p(t)/q(t)\propto t^2$. This gives us
\begin{equation}
    \begin{split}
    L(\delta) &= \E_{t\sim p(t)}\big[L(\delta;t)\big]
    = \int_\varepsilon^Tp(t)\,L(\delta;t)\,\dd t\\
    &\propto\int^T_\varepsilon q(t)\,t^2\,L(\delta;t)\,\dd t
    =\E_{t\sim q(t)} \big[ t^2\,L(\delta;t)\big],
    \end{split}
\end{equation}
which we use as our objective for the experiments in \S\ref{sec:experiments}.

\section{Extension to TrigFlow}
\label{sec:stad-extensions} 
Flow-map models \citep{sabour2025alignflowscalingcontinuoustime} have been introduced to unify recent frameworks, such as consistency models \citep{song2023consistency, song2024improved, lu2025simplifying} and mean flow \citep{geng2025mean}, that were developed for faster sampling under diffusion and flow-based models. Of these formulations, the TrigFlow continuous-time consistency model \citep{lu2025simplifying} is a particularly natural fit for \textit{StAD}.

\citet{lu2025simplifying} introduce TrigFlow, that optimises the objective
\begin{equation}
    \label{eq:trigflow-loss}
    \mathcal{L}(\theta) = \E_{t}\,\E_{\xx_\varepsilon}\E_{\xx_t|\xx_\varepsilon}\Big[\big|\big|\sigma_d\vv_t(\xx_t;\theta) - \mathbf{u}_t\big|\big|^2\Big]
\end{equation}
where $\sigma_d$ is the variance in the data distribution, and $\vv_t(\xx_t;\theta)$ is a model vector field with parameters $\theta$. Here, the noising schedule is defined by $\xx_t=\cos(t)\xx_\varepsilon+\sin(t)\mathbf{z}$ with $p(\mathbf{z})=N(0,\sigma^2_d\mathbf{I})$, equivalent to the flow-matching noise schedule \eqref{eq:noisingschedule} with $\alpha_t = \cos (t)$ and $\sigma_t = \sigma_d\sin (t)$. Thus the flow matching objective's target can be defined as $\mathbf{u}_t \coloneqq -\sin (t)\xx_\varepsilon + \cos (t) \mathbf{z}$, giving us
\begin{equation}
    \xx_\varepsilon = \cos (t) \xx_t - \sin (t) \mathbf{u}_t.
\end{equation}
The conditional expectation as shown by \citet{zhou2025scoredistillationflowmatching} is $\E[\xx_\varepsilon | \xx_t] = \cos (t) \xx_t - \sin (t) \E[\mathbf{u}_t | \xx_t]$, under the flow-matching regression objective  $\vv_t(\xx_t) \approx \E[\mathbf{u}_t | \xx_t]$. Now, using \eqref{eq:gaussian-score}, we can derive the score for TrigFlow models under Gaussian assumptions as:
\begin{equation}
    \score_t(\xx_t) = - \frac{\xx_t - \cos (t)\E[\xx_\epsilon | \xx_t]}{\sin^2(t)}
\end{equation} 
Using the trigonometric path relation gives us:
\begin{equation}
    \begin{split}
    \score_t(\xx_t) &\approx -\frac{\xx_t -\cos (t)\,\big( \cos (t)\xx_t - \sin (t) \vv_t(\xx_t)\big)}{\sin^2(t)} \\
     &\approx - \big(\xx_t + \cot (t)\vv_t(\xx_t)\big)
    \end{split}
\end{equation}
This allows us to define the baseline,
\begin{equation}
    b_t(\xx_t) = \inner{\vv_t(\xx_t)}{\xx_t} + \cot (t)||\vv_t(\xx_t)||^2
\end{equation}
and the residual,
\begin{equation}
    r_t(\xx_t) = \inner{\nabla}{\vv(\xx_t)} - \inner{\vv_t(\xx_t)}{\xx_t} - \cot (t)||\vv_t(\xx_t)||^2.
\end{equation}
Similarly to flow-matching (\S\ref{sec:stad-flow-matching}), the rest of the \textit{StAD} objective remains the same. This allows us to approximate the divergence and compute log-likelihood cheaply for consistency models. 

\section{Experiment details}
\label{sec: detailedexperiments}
\subsection{Hardware}
\label{sec:hardware}
All of the experiments conducted here were run on NVIDIA-A100 GPUs with 40GiB VRAM and 1TB of RAM on 48 core nodes. Each node utilised a $2 \times 24$ Core AMD EPYC 7402 2.8~GHz processor.

\subsection{Software}
We use the \texttt{torchdiffeq} \citep{chen2018node} library for all the experiments, with the  \texttt{dopri5} adaptive Runge--Kutta ODE solver \citep{dormand80rk}. We set the absolute and relative tolerances to $10^{-5}$ for our solvers for \textit{StAD}, the stochastic estimators, and the exact trace. All code is implemented in \texttt{PyTorch} \citep{paszke2019pytorch}.

\subsection{\textit{StAD} implementation}
We describe the \textit{StAD} distillation procedure for diffusion in \cref{alg:stad-train}, for the flow-matching case the algorithm stays the same, but we build the cache not from SDE marginals, but by evaluating the teacher velocity at $\xx_t$.
\begin{algorithm}
\caption{Cached Stein distillation for conditional diffusion (\textit{StAD} implementation)}
\label{alg:stad-train}
\begin{algorithmic}[1]
\REQUIRE data samples and contexts $\{(\xx_\varepsilon,c)\}$; teacher PF-ODE drift $\vv_\theta(\xx_t;t,c)$; scalar head $\delta_\phi(\xx_t;t,c)$; SDE conditional mean, variance $\nu(t)$, $\eta^2(t)$.
\REQUIRE proposal $q(t)$; target $p(t)=U(\varepsilon,t)$; cache size $M$; steps $S$; batch size $B$; rebuild period $K$; regularization $\kappa_R(||\xx_t||)$; $L^2$ penalty $l$.
\STATE build cache $\mathcal{C}=\{(\xx_t,t,\vv_t,c)\}_{m=1}^M$:
\FOR{$m=1$\dots$M$}
  \STATE sample $(\xx_\varepsilon,c)\sim\text{data}$
  \STATE normalize $\xx_\varepsilon\leftarrow (\xx_\varepsilon-\text{shift})/\text{scale}$; $c$ if used
  \STATE sample $t\sim q(t)$, $\mathbf{z}\sim\mathcal{N}(0,\mathbf{I})$
  \STATE $\xx_t \leftarrow \nu_t(\xx_\varepsilon) + \eta(t) \mathbf{z}$
  \STATE $\vv_t \leftarrow \vv_\theta(\xx_t;t,c)$
  \STATE store $(\xx_t,t,\vv_t,c)$ in $\mathcal{C}$
\ENDFOR
\FOR{$s=1$\dots$S$}
  \STATE sample $\{(\xx_i,t_i,\vv_i,c_i)\}_{i=1}^B$ uniformly from $\mathcal{C}$
  \STATE $w_i \leftarrow p(t_i)/q(t_i)$
  \STATE $\lambda_i \leftarrow \kappa_R(||\xx_i||)$
  \STATE $\delta_i \leftarrow \lambda_i\delta_\phi(\xx_i;t_i,c_i)$
  \STATE $\mathbf{g}_i \leftarrow \nabla_{\xx_i}\,\delta_i$
  \STATE $\mathcal{L} \leftarrow \frac1B\sum_{i=1}^B w_i\Big(\delta_i^2 + 2\langle\mathbf{g}_i,\vv_i\rangle + l||\mathbf{g}_i||^2\Big)$
  \STATE update $\phi$ using $\nabla_\phi \mathcal{L}$
  \IF{$K>0$ \AND $s \bmod K=0$} \STATE rebuild $\mathcal{C}$ as above \ENDIF
\ENDFOR
\end{algorithmic}
\end{algorithm}
\subsection{$\delta$ regularisation}
For most of our experiments we set $R$ as a high quantile of $\lvert\lvert\xx\rvert\rvert$. For the astrophysical flux, \texttt{ImageNet-32x32} and \texttt{CIFAR-10} experiments, we set $R$ to the 99.5th percentile of $\lvert\lvert\xx\rvert\rvert$ . We generally follow the cosine function defined in \S\ref{sec:regularisation}. However we can also make harder constraints on $\delta$ by using a bump function, where,
$\kappa_R \propto \exp \big (1/\lvert \vert \xx \rvert \rvert^2) \text{ if } R < \lvert \vert \xx \rvert \rvert <2R$.

\subsection{Hyperparameters}
\subsubsection{Astrophysical fluxes}
\paragraph{Training}
We describe the hyperparameters used to train the diffusion model to predict astrophysical flux uncertanties in \cref{tab:cosmostraininghp}.
\begin{table}
    \centering
    \caption{Diffusion model hyperparameters for flux uncertainty prediction in the \texttt{COSMOS2020} dataset.}
    \label{tab:cosmostraininghp}
    \small
    \begin{tabular}{lcccc}
    \toprule
    \textbf{SDE} &
    \textbf{$\beta_\text{max}$} &
    \textbf{epochs} &
    \textbf{LR} &
    \textbf{optimiser}\\
    \midrule
    VP & $20$ & $1000$  & step; 1e-4 to 1e-5 & $\texttt{Adam}$ \\ 
    \bottomrule
    \end{tabular}
\end{table}
We use tanh activation functions for the MLP and employ a step scheduler with the step size of half the learning rate, alongside with a heating step schedule of $2 \times$ the batch size, starting with 64.

\paragraph{Distilling}
We describe the hyperparameters used to distill the astrophysical model in \cref{tab:cosmosdistillinghp}. We use a cosine scheduler for the learning rate. Each epoch is extremely efficient; we observe wall-clock of $\sim$ 1.5 hours to achieve the accuracy showed in \cref{img:astroresiduals} on the hardware specified in Appendix \ref{sec:hardware}.
\begin{table}
    \centering
    \caption{MLP head, $\delta$, for the VP-SDE diffusion model trained on \texttt{COSMOS2020}. We set $R$ to the $P_R$th percentile of $||\xx||.$ }
    \label{tab:cosmosdistillinghp}
    \small
    \begin{tabular}{lcccc}
    \toprule
    \textbf{MLP} &
    \textbf{$P_R$} &
    \textbf{epochs} &
    \textbf{LR} &
    \textbf{optimiser}\\
    \midrule
    $512 \times 3$ & $99.5\%$ & $1\times10^5$  & 1e-4 to 1e-8 & $\texttt{Adam}$ \\ 
    \bottomrule
    \end{tabular}
\end{table}
\subsubsection{ImageNet-$32\times$32 and CIFAR-10}
\paragraph{Training}
We train a U-Net architecture, whose hyperparameters are listed in \cref{tab:CIFARhp} and \cref{tab:ImageNethp}, respectively for \texttt{CIFAR} and \texttt{ImageNet-32x32} models. 
\begin{table}
\centering
\caption{Training hyperparameters for \texttt{CIFAR-10}.}
\label{tab:CIFARhp}
\begin{tabular}{ll}
\toprule
\textbf{hyperparameter} & \textbf{value} \\
\midrule
SDE & VP-SDE \\
loss & denoising score matching \\
base channels & 128 \\
channel multipliers & (1, 2, 2, 2) \\
ResBlocks per level & 2 \\
attention resolutions & 16, 8 \\
attention heads & 4 \\
optimizer & \texttt{AdamW} \\
learning rate & $2 \times 10^{-4}$ \\
batch size & 256 \\
epochs & 400 \\
gradient clipping & 1.0 \\
EMA decay & 0.999 \\
sampling steps & 200 \\
\bottomrule
\end{tabular}
\end{table}
\begin{table}
\centering
\caption{Training hyperparameters for \texttt{ImageNet-32x32}.}
\label{tab:ImageNethp}
\begin{tabular}{ll}
\toprule
\textbf{hyperparameter} & \textbf{value} \\
\midrule
SDE & VP-SDE \\
loss & denoising score matching \\
base channels & 128 \\
channel multipliers & (1, 2, 2, 2) \\
ResBlocks per level & 2 \\
attention resolutions & 16, 8 \\
attention heads & 4 \\
optimizer & \texttt{AdamW} \\
learning rate & $2 \times 10^{-4}$ \\
batch size & 256 \\
epochs & 200 \\
gradient clipping & 1.0 \\
EMA decay & 0.999 \\
sampling steps & 200 \\
\bottomrule
\end{tabular}
\end{table}

\paragraph{Distilling}
We distill the PF-ODEs learned by these models, using a compact CNN architecture, with FiLM conditioning for injecting time into data. Hyperparameters and further architectural details for the distillation head are described in \cref{tab:CIFARdistillhp} for \texttt{CIFAR-10} and in \cref{tab:ImageNetdistillhp} for \texttt{ImageNet-32x32}.
\begin{table}
\centering
\caption{Distillation hyperparameters for \texttt{CIFAR-10} $\delta$-head.}
\label{tab:CIFARdistillhp}
\begin{tabular}{ll}
\toprule
\textbf{hyperparameter} & \textbf{value} \\
\midrule
\multicolumn{2}{l}{\textit{architecture}} \\
head type & FiLM-CNN \\
base channels & 64 \\
time embedding dim & 256 \\
group norm groups & 16 \\
time input & $\log t$ \\
downsampling & $32 \to 16 \to 8$ (stride-2 convs) \\
output & global pool $\to$ MLP $\to$ scalar \\
\midrule
\multicolumn{2}{l}{\textit{training}} \\
optimizer & \texttt{AdamW} \\
learning rate & $1 \times 10^{-4}$ \\
training steps & 600,000 \\
batch size & 2,048 \\
gradient clipping & 1.0 \\
\midrule
\multicolumn{2}{l}{\textit{caching}} \\
cache samples & 262,144 \\
cache refresh interval & 2,000 steps \\
time sampler & uniform \\
\midrule
\multicolumn{2}{l}{\textit{cutoff}} \\
cutoff mode & cosine \\
$P_R$ & 99.5\% \\
\bottomrule
\end{tabular}
\end{table}
\begin{table}
\centering
\caption{Distillation hyperparameters for \texttt{ImageNet-32x32}.}
\label{tab:ImageNetdistillhp}
\begin{tabular}{ll}
\toprule
\textbf{hyperparameter} & \textbf{value} \\
\midrule
\multicolumn{2}{l}{\textit{architecture}} \\
head type & FiLM-CNN \\
base channels & 64 \\
time embedding dim & 256 \\
group norm groups & 16 \\
time input & $\log t$ \\
downsampling & $32 \to 16 \to 8$ (stride-2 convs) \\
output & global pool $\to$ MLP $\to$ scalar \\
\midrule
\multicolumn{2}{l}{\textit{training}} \\
optimizer & \texttt{AdamW} \\
learning rate & $1 \times 10^{-4}$ \\
training steps & 600,000 \\
batch size & 2,048 \\
gradient clipping & 1.0 \\
directional method & exact gradient (autograd) \\
\midrule
\multicolumn{2}{l}{\textit{caching}} \\
cache samples & 262,144 \\
cache refresh interval & 2,000 steps \\
time sampler & uniform \\
\midrule
\multicolumn{2}{l}{\textit{cutoff}} \\
cutoff mode & cosine \\
$P_R$ & 99.5\% \\
\bottomrule
\end{tabular}
\end{table}

\subsection{Stochastic trace estimators}
We use Rademacher distributed test vectors for all our Hutchinson tests. We base our implementation of Hutch++, including caching of the $QR$ decomposition, on \citet{liu2025trace}. We update the cached $\mathbf{Q}$ matrices once per Runge--Kutta step (i.e.\ every sixth function evaluation for the \texttt{dopri5} solver). Our implementation of XTrace is based in the efficient algorithm described in the supplementary material of \citet{Epperly_2024}. 

\begin{figure}
  \begin{center}
    \centerline{\includegraphics[width=\linewidth]{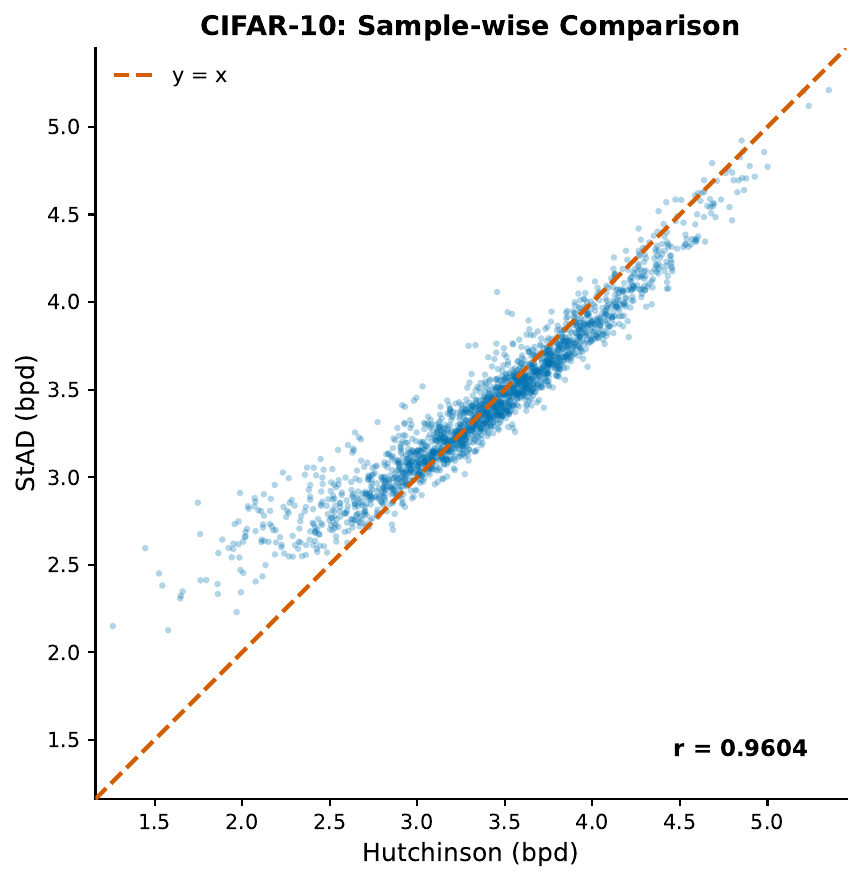}}
    \caption{
    Image-wise NLL comparison (\textit{StAD} vs.\ Hutchinson with 2 probes) for
2048 images from \texttt{CIFAR-10}.}
    \label{img:CIFAR_scatter}
  \end{center}
\end{figure}
\begin{figure*}
  \begin{center}
    \centerline{\includegraphics[width=\linewidth]{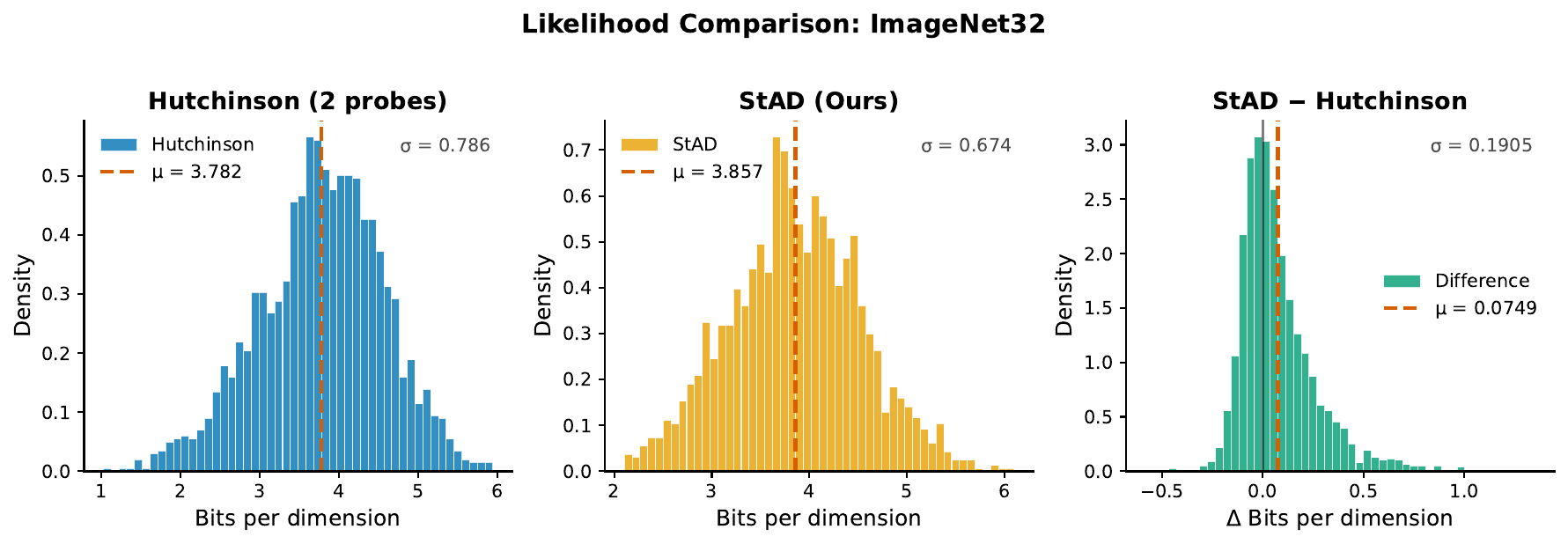}}
    \caption{
        NLL histograms for 2048 images from \texttt{ImageNet-32x32}. \textit{Left}: Using the Hutchinson algorithm with 2 probes to estimate the NLL. 
    \textit{Centre}: Using \textit{StAD}. \textit{Right}: Difference between \textit{StAD} and Hutchinson: $\text{NLL}[\textit{StAD}] - \text{NLL}[\text{Hutchinson}]$.}
    \label{img:ImageNet_comparision}
  \end{center}
\end{figure*}

\section{Further ImageNet and CIFAR-10 results}
\label{sec:image-results-extra}
In \cref{img:CIFAR_scatter} we show the correlation between NLLs from the Hutchinson (2-probe) estimator and \textit{StAD} for 2048 samples from \texttt{CIFAR-10}. In \cref{img:ImageNet_comparision} we compare NLL histograms from \textit{StAD} and Hutchinson for \texttt{ImageNet-32x32}.

\section{CIFAR-10 and ImageNet samples}
\label{sec:horse}
\begin{figure*}
  \begin{center}
    \centerline{\includegraphics[width=\linewidth]{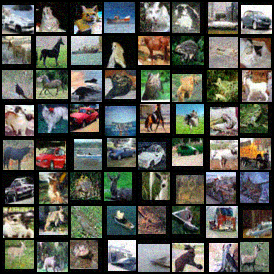}}
    \caption{
    Unconditional \texttt{CIFAR-10} samples from a VP-SDE diffusion model trained with a U-Net architecture.}
    \label{img:CIFAR_samples}
  \end{center}
\end{figure*}

\begin{figure*}
  \begin{center}
    \centerline{\includegraphics[width=\linewidth]{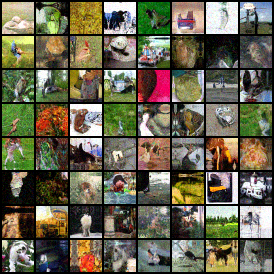}}
    \caption{
    \texttt{ImageNet-32x32} samples from a VP-SDE diffusion model trained with a U-Net architecture.}
    \label{img:ImageNet_samples}
  \end{center}
\end{figure*}

In \cref{img:CIFAR_samples} and \cref{img:ImageNet_samples}, we show example images generated by our teacher diffusion models (in both cases based on a VP-SDE and U-Net score network) for the \texttt{CIFAR-10} and \texttt{ImageNet-32x32} training datasets.




\end{document}